\documentclass[preprint,3p]{elsarticle}
\usepackage{longtable}
\usepackage{adjustbox}
\usepackage{lscape} 
\usepackage{lineno,hyperref}
\usepackage{graphicx} %package to manage images
\usepackage[english]{babel}
\usepackage[autostyle]{csquotes}
\usepackage{soul,color}
\usepackage{multirow}

\journal{Simulation Modelling Practice and Theory}

\begin{document}

\begin{frontmatter}
\title{Application Specific Drone Simulators: Recent Advances and Challenges}

%\author{Aakif Mairaj}
\author[1]{Aakif Mairaj}%\fnref}{fn1}}
\ead{Aakif.Mairaj@UToledo.Edu}
\author[2]{Asif I. Baba}%\fnref}%{fn1}}
\ead{ababa@tuskegee.edu}
\author[1]{Ahmad Y. Javaid\corref{cor1}}%\fnref{fn2}}
\ead{Ahmad.Javaid@UToledo.Edu}


\cortext[cor1]{Corresponding author}

\address[1]{2801 W. Bancroft St MS 308, The University of Toledo, Toledo, 43606, USA}
\address[2]{1200 W Montgomery Rd, Tuskegee University, Tuskegee, AL 36088, USA}
%\address[focal]{The University of Toledo, Toledo, USA}

\begin{abstract}
Over the past two decades, Unmanned Aerial Vehicles (UAVs), more commonly known as drones, have gained a lot of attention, and are rapidly becoming ubiquitous because of their diverse applications such as surveillance, disaster management, pollution monitoring, film-making, and military reconnaissance. However, incidents such as fatal system failures, malicious attacks, and disastrous misuses have raised concerns in the recent past. Security and viability concerns in drone-based applications are growing at an alarming rate. Besides, UAV networks (UAVNets) are distinctive from other ad-hoc networks. Therefore, it is necessary to address these issues to ensure proper functioning of these UAVs while keeping their uniqueness in mind. Furthermore, adequate security and functionality require the consideration of many parameters that may include an accurate cognizance of the working mechanism of vehicles, geographical and weather conditions, and UAVNet communication. This is achievable by creating a simulator that includes these aspects. A performance evaluation through relevant drone simulator becomes indispensable procedure to test features, configurations, and designs to demonstrate superiority to comparative schemes and suitability. Thus, it becomes of paramount importance to establish the credibility of simulation results by investigating the merits and limitations of each simulator prior to selection. Based on this motivation, we present a comprehensive survey of current drone simulators. In addition, open research issues and research challenges are discussed and presented.

\end{abstract}

\begin{keyword}
\texttt Unmanned Aerial Vehicles \sep UAVs \sep FANETs \sep UAV Simulator

%\MSC[2018] 00-01\sep  99-00
\end{keyword}

\end{frontmatter}

%\linenumbers

\section{Introduction}

Although the correct date of the first successful UAV flight is not clear, some circles claim that the history of UAVs dates back to 1849, when the Austrians launched 200 pilot-less balloons against the city of Venice~\cite{2185196}. Gradually, the use of drones diversified in many areas of aviation, and currently, UAVs are gaining a considerable acceptance because of their applications in a multitude of fields~\cite{Vijayweb,javaid2015single,schiavullo2018ehang,blackmore2014farming,TheFutur99:online}. The drones are believed to perform the tasks which are dangerous, dull, dirty or dumb~\cite{JAST}. In addition to their use in military missions~\cite{Military23:online}, disaster management~\cite{HowDrone4:online}, geographical monitoring~\cite{TheUseof74:online}, journalism \cite{Dronesan5:online}, crop monitoring \cite{Agricult98:online}, precision farming \cite{Dronesan12:online}, archaeology, and pizza delivery \cite{DronesTr80:online, HBOUsedP79:online}; in future they are expected to be used for taxi services too \cite{Dronetax24:online}. Utilizing UAVs for such applications is relatively inexpensive compared to manned aerial vehicles~\cite{Vogeltanz20165}. In many applications, multiple UAVs are required for carrying out successful missions collaboratively. In such applications, one of the key features is the network coordination between UAVs and equally important is their security~\cite{shrit2017new}. However, dealing with UAV-Networks (UAVNets) is challenging. Recent studies have shown the possibility of attacks on the control and data communication streams other than jamming, denial of service, and spoofing attacks~\cite{8088163,hartmann2013vulnerability}. UAVNets are different from traditional wireless networks in terms of mobility, processing power, localization, power consumption, and communication protocols. UAVs come in diverse shapes and sizes, and their aerodynamic models vary with their types. A flaw in aerodynamics behavior of UAV puts the cost, confidentiality, and the success of a UAV mission under threat, thus making the design and functionality of UAVs challenging and costly. Therefore, it is crucial to use a testbed or a simulator that conforms to the viability standards of UAVs before finalizing a design or making them airborne. In this paper, we will use word UAVs and drones interchangeably.

One of the most important ideas driving the use of drones is the concept of the Internet of Things (IoT)~\cite{ColinSnow}. It fits International Telecommunication Union's (ITU) definition of the IoT: connecting anything, anytime, anyplace, for anyone \cite{DBLP:journals/corr/AbdmeziemT14}. The technology that delivers drones with IoT capabilities exists today. The combined technologies are aimed at delivering near real-time analytics on spatio-temporal data captured over a long time. With the increasing number of networking devices and several ubiquitous technologies coming together, the challenges are growing. Some of them include security, Quality of Service (QoS), and development of protocols for different layers. To recognize such issues, some simulators and emulators have been developed~\cite{6bestIoT0:online,SimpleSo75:online}. The simulator designers and developers should keep in mind the aerodynamic nature, 3-D flight, and network communication of drones while building a simulator. In many applications, a multi-UAV system is preferable over a single powerful drone for several possible reasons. First, the ability of a single UAV to sense and actuate at a time is limited while as a swarm of UAVs can simultaneously collect information from multiple locations and exploit the information derived from multiple disparate points to build models that can be used to make decisions. Additionally, multiple UAVs can apply forces simultaneously at different locations to perform actions that could be very difficult for a single UAV. % (Note: What force you are talking about? Akif.)
Second, efficiency, cost, the execution time of missions such as exploration and searching for targets can be drastically reduced by UAV swarms. For certain tasks, using a network of low-cost drones instead of single UAV can also be cost-effective. In terms of {\bf reliability}, the multi-UAV approach leads to redundant solutions offering greater fault tolerance and flexibility including re-configurability in case of failures of individual vehicles.

Security of the data and control links within these networks are critical for the confidentiality and success of a mission. Furthermore, the dynamic nature of UAVs pose some additional challenges, e.g., if a single node goes down, self-organization is required, and if not addressed carefully, it may affect the availability~\cite{gupta2016survey}. Difficulties in self-organization make this an interesting research problem which could be further explored through simulation.

\subsection{UAV Simulator Requirements}
There are three main functions of UAV simulators: evaluating new technology, low-cost training, and research \& development (R\&D). Modeling and simulating a UAV is not an easy task, it requires the calculation of many parameters. %For example,
Unlike the mobility of Vehicular Ad-hoc Networks (VANETs), the movement of a UAV is not confined to 2D space~\cite{article19}. In 3D mobility model of UAVs, calculation of various aerodynamic coefficients add more challenge~\cite{Vogeltanz20165}. UAV simulators are used to create complex mathematical models that resemble the real physical system to a higher degree and reduce the experimentation time and cost~\cite{article19}. There are many commonalities between UAV simulators and traditional flight simulators, e.g., both find applications in the aircraft and flight mechanics design, and enable the operators to train for hazardous situations~\cite{8088163}. Henceforth, we are using a term UAV simulator and flight simulator interchangeably in this paper.

A good UAV simulator requires the following characteristics: easy to use, should run on general purpose hardware, less complex, and easy GUI~\cite{article19}. A detailed description of some of the important requirements for a good UAV simulator is provided below \cite{Vogeltanz20165,allerton2009principles}:

\begin{enumerate}

\item {\bf Flight dynamics model}: The airfoil of an aircraft is capable of generating sufficient lift for a flight. Moreover, these aircraft are user-controlled, either by onboard auto-pilots or remote computers. Understanding the aerodynamic forces produced during flight is crucial for numerous reasons. First and foremost, the air-frame has to bear the applied forces. Second, an understanding of the aerodynamic forces results in efficient aircraft design. Third, the aircraft is controlled by a pilot, suggesting an understanding of the dynamics of the aerial vehicle. Fortunately, for the flight simulator designer, many of the existing equations related to the aerodynamics are well understood and can be applied to the dynamics modeling of a drone.

\item {\bf System model}: Concurrently, the credibility of a drone simulator relies on an accurate mathematical model of the UAV system. Developed models are either based on an understanding of the physics of the overall system or derived from the reliable real-world data obtained from UAVs. The aim is to mimic the behavior of a real UAV in a simulator so that the set of I/O relationships is identical for both the vehicle and its simulated model, i.e., if a particular set of inputs to the drone produces a specific set of outputs, the same set of inputs would produce identical outputs in the simulator.

\item {\bf Graphical model}: Aircraft heads-up displays (HUD) form a specialized subset and area of computer graphics research. The graphical models of a UAV rely upon few colors, no shading, vector graphics, alphanumeric text, and few related symbols. The displays must update at least 20 times/s, and the display functions are applied using 2D graphics procedures. Mostly, the display is considered on a Cartesian coordinate system where the x-axis is horizontal, and the y-axis is vertical. At times, there are differences in the convention based on the landscape or portrait display mode. Such conventions are generally defined in a graphics library, allowing the user to develop graphics software independent of any hardware limitations, as shown in Figure~\ref{fig:graphics}.

\begin{figure}[h]
\centering
\includegraphics[width=0.6\textwidth]{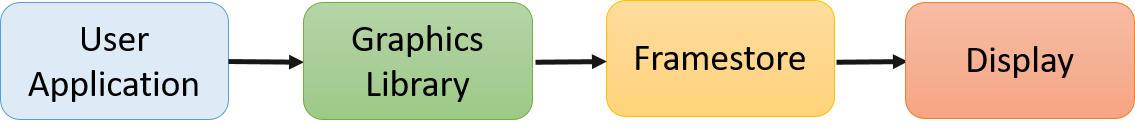}
\caption{The graphics pipeline}
\label{fig:graphics}
\end{figure}

\item {\bf Control system}: Flight simulators treat control systems as mathematical transfer functions, and hence the simulator designer has to incorporate the transfer functions in algorithmic form to make sure that the I/O relationship of a system is modeled correctly. The system engineers consider the time response, the frequency response and the transfer function as equivalent and interchangeable~\cite{allerton2009principles}.

\item {\bf Flight route identification}: The navigation system in an aircraft enables the users to locate current and future positions. Various other subsystems work in conjunction with the navigation system to enable features such as flight route identification, collision avoidance, path planning, and air-corridor assignment. However, in the flight simulator, there are no signals or sensors, and the simulator designers have to face tough challenges. The designers fabricate a simulator of each subsystem that replicates the behavior of the real aircraft subsystem. The designer makes sure that all operational modes are fully replicated and that there is no difference between the simulated component and the actual component.

\item{\bf UAV types and models}: A good simulator should also support a variety of drone types and models. Although there is no particular standard for classifying UAVs, some experts categorize them based on parameters such as size, range, and configuration. In Table~\ref{classification}, an overview of different kinds of UAVs categorized based on size~\cite{ColinSnow,article15}, endurance, and configuration are provided. Their applications might vary with the size, e.g., the larger UAVs find their applications in military and combat operation while the smaller UAVs are used for surveillance, academic research, and entertainment. Similarly, the UAVs with greater endurance find their applications in surveillance missions, whereas the smaller UAVs work well for traffic monitoring, and entertainment. On the basis of configuration, drones can be classified as fixed wing, rotary wing, and flapping wing - all differ from each other because of their aerodynamic design and functionality. Fixed wing drones are typically larger, and they fly over a considerable distance. Rotary wing drones are among the most commonly used UAVs. Flapping wing drones are bio-inspired, as they emulate the flying pattern of birds and insects, and are still mostly under R\&D.

\item{\bf Application-specific requirements}: There might be additional application-specific requirements for UAV simulators. A good UAVNet simulator might require the augmentation with communication aspect of drones, where each UAV is treated as a node in the network, and communication is established using UAVNet-specific networking protocols. Similarly, a simulator being used for cybersecurity research might need additional modules on possible attacks. Simulators might require a module for detailed data or result analysis. The speed of the simulator should be adjustable to allow the users to extract essential details~\cite{javaid2015single}. Furthermore, the simulations should be executable in different scenarios with different mobility models~\cite{javaid2013uavsim}. Additionally, several assumptions are made for the ease of modeling such as aircraft being rigid, the earth being the inertial reference frame, and constant aircraft mass throughout the simulation~\cite{Vogeltanz20165}.

\end{enumerate}

\begin{figure}[h]
\centering
\begin{minipage}[b]{0.45\linewidth}
\centering
\includegraphics[width=1.1\textwidth]{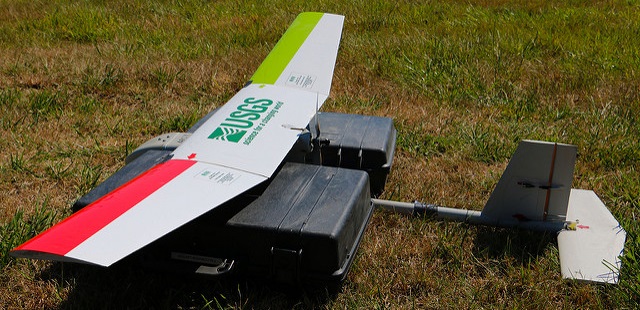}
\caption{Fixed-wing UAV}
\label{FixedWing}
\end{minipage}
\hspace{1.0cm}
\begin{minipage}[b]{0.45\linewidth}
\centering
\includegraphics[width=0.935\textwidth]{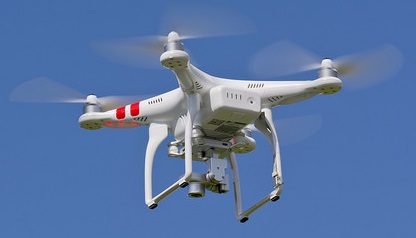}
\caption{Rotary-wing UAV DJI Phantom 2 Vision+}
\label{DJI}
\end{minipage}
\end{figure}

\begin{longtable}{p{55pt} p{220pt} p{150pt}}
\caption{Classification of UAVs~\cite{ColinSnow,article15}}
\label{fig:figure4}\\
\hline
\textbf{Category} & \textbf{Sub type(s)} & \textbf{Properties}\\
\hline
\multirow{5}{*}{Size} &\textbf{Large- scale UAVs}: Fly either autonomously or through an operator. 
& Size (10m), Weight (-) \\\cline{2-3}
&\textbf{Medium-scale UAVs:} Too heavy to be carried by a person; however, lighter than the large-scale UAVs. & Size (5m), Weight (-) \\\cline{2-3}
&\textbf{Small-scale UAVs:} Mostly fixed-wing. & Size (50cm-2m), Weight (2-10kg) \\\cline{2-3}
&\textbf{Mini UAVs:} Most commonly used UAVs, cheap, easy to maintain and smaller in size. &  Size (1-1.4m), Weight (-) \\\cline{2-3}
&\textbf{Micro/Nano UAVs:} Used for surveillance and academic research. & Size (-), Weight (100g) \\%\cline{2-3}
\hline
\multirow{4}{*}{Endurance} & \textbf{Very low cost and close range UAVs:} Include Mini, Nano, and Micro UAVs. & Endurance (0.5-1 hour), Range (5km)  \\ \cline{2-3}
&\textbf{Close Range UAVs:} Primarily used in surveillance, recon, traffic monitoring, etc.  & Endurance (1-6 hours), Range (100 km) \\ \cline{2-3}
&\textbf{Mid Range UAVs:} Very high speed, used for surveillance, gathering meteorological data.    & Endurance (24 hours), Range (650km) \\ \cline{2-3}
&\textbf{Long-Endurance UAVs:} Used for long-range surveillance missions. & Endurance (36 hours) and Maximum Height (9100m) \\ \hline
\multirow{3}{*}{Configuration} & \textbf{Fixed wing:} Capable of flying over longer distances, carrying a heavy payload with low power consumption. Modeling and structure is simpler than rotary wing (e.g., Figure~\ref{FixedWing}). & Need a runway for takeoff and landing. \\ \cline{2-3}
& \textbf{Rotary wing:} Quite maneuverable, have less endurance, and noisy. Further classified into the single rotor, quadcopter, hexa-copter, and octa-copter. (e.g., Figure~\ref{DJI}). & Biggest advantage is the ability to land and take off vertically. \\ \cline{2-3}
& \textbf{Flapping wing:} Flight dynamics is bio-inspired from insects and birds, and is more complicated than other UAVs because of the nature of flight. & Bridges the gap between fixed wing and rotary wing flights. Less noisy compared to rotary wing UAVs. \\ \cline{2-3}
\hline
\label{classification}
\end{longtable}

The process of developing a drone simulator is interdisciplinary. It involves collaborative efforts from different disciplines, therefore making it a fascinating subject, which requires a comprehensive explanation to make it understandable to the participants. Following are the primary contributions of this work: 

\begin{enumerate}
\item Discussion on the evolution of drone simulators from the historical perspective, providing the readers with a picture of the origins of the UAV simulation industry and its growth in the recent past.
\item In-depth analysis of the drone simulator requirements that would enable the readers to understand the discussed simulators from various perspectives, i.e., aerodynamics, game engineering, network engineering, and software development.
\item Detailed discussion on popular simulators of different genres along with their comparison, summarized in a tabular form.
\item Elaborated discussion on current challenges and future research directions that touches upon the recent attacks on UAVNets, swarming technology, and bio-inspired drones.
\end{enumerate}

\section{Background and Related work}
The first flight simulator, independent of wind, was used in 1910 for the training of pilots. It consisted of two halves of a barrel, mounted and operated manually by a pilot seated in the upper half barrel to control pitch and roll of an aircraft, as shown in Figure~\ref{fig:figure4}. The pilots were required to align the reference bar of the simulator with the horizon~\cite{article20,article21}.

\begin{figure}[h]
\centering
\includegraphics[width=9cm]{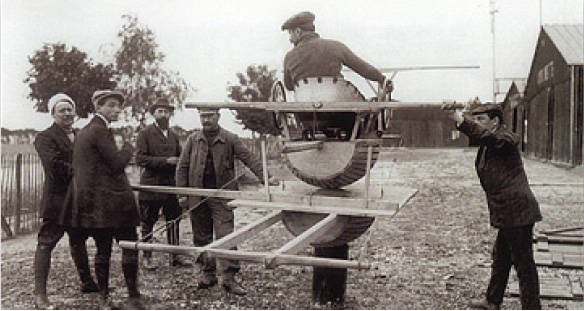}
\caption{Early Manual Flight Training Devices~\cite{article21}}

\end{figure}

Further, World War I encouraged the development of better flight training machines for aviators. These simulators were installed with control and electrical recording apparatus. Later, the development of these flight trainers moved towards a stage where installed electrical and mechanical equipment was meant to provide real aircraft responses to the control inputs of the pilot. Edwin Link developed one of the most popular and successful flight simulating device in Binghamton, New York, USA between 1927-1930. It was considered to be the most efficient and novel training device of that time and became the most popular and preferred pilot trainer during World War II. Nearly 10,000 of these devices were made to train 500,000 new pilots from different countries and the \textit{Link Trainer} was used by almost all the US Air Force pilots. Link trainer refers to the series of flight simulators that were used between the 1930-50s~\cite{article20,article21,article22,article23,article24}.

During World War II, electronic simulators started getting attention as computers were now capable of solving the flight dynamics equations. Curtiss Wright designed the first complete simulator for an airline, known as \textit{Pan America} during the 1950s. US Airlines purchased this simulator for \$3 million and is considered to be among the first modern flight simulators as it includes aircraft visuals and sounds. During the 1960s, simulators started including motions across pitch, roll and yaw axes. In the 1970s, computer graphics simulators were introduced that were very basic, with a single mobile white dot on a black landscape. Gradually, the displays evolved into more detailed and comprehensive interactive screens. One of the biggest achievements within this industry was the Microsoft Flight Simulator~\cite{Cumming}. It began as a collection of 3D flight simulation graphics articles authored by Bruce Artwick in 1976. Its first version, Flight simulator 1.0, was released in 1982. The latest and most advanced version is Flight Simulator X and includes over 24,000 airports with accurate and real-time weather conditions~\cite{javaid2013uavsim, article20,article21,article22,article23,article24,Cumming,article26,article27,article28}.

Besides, numerous recent attacks on UAVs and newly introduced Flying Ad-Hoc Network (FANET) protocols have caused industrialists, researchers and hobbyists to simulate and test various UAV models, their networks, and flight dynamics of UAV flights. Some of these simulators and testbeds have been created using the OMNeT++ network simulator~\cite{javaid2013uavsim, marconato2017avens}, that enables UAVNet simulation. Several of the popular networking simulators do not model multi-UAV systems. To the best of our knowledge, there are no built-in models for UAV communication and infrastructure in simulators such as OPNET and NS-2. These simulators cannot model the communication in FANETs [32,33]. Most 3D simulators are used for gaming or design engineering. There are two main types of 3D drone simulators: first, to help drone-enthusiasts practice First Person View (FPV), and second, to refine the handling of camera drones to capture good quality videos and photos~\cite{tr}. UAV simulators can be classified on the basis of whether they (i) will work under outdoor or indoor conditions, (ii) are for research or entertainment, (iii) consider single, or a network of UAVs, and (iv) are purely software/hardware or hybrids~\cite{jahan2015gnssim,javaid2015cyber}.

\section{UAV-Simulators and their Applications}

Drone enthusiasts, design engineers, and researchers require simulators for numerous reasons such as practicing and improving videography skills for FPV, aerial or Line of Sight (LOS) view. Additionally, based on the application domain, graphic detail in a simulator is essential for a realistic experience and technically sound aerodynamic design. The correct aerodynamic design of a UAV is vital for all different shapes, sizes and configurations, otherwise, it may lead to a crash. Therefore, it is necessary to model it correctly within a simulator. Besides, it is no longer critical to building a simulator from scratch because researchers can choose from several existing game engines, open source (O/S) networking simulators, and commercial simulators to develop their solution.

\subsection{HEXAGON Flight Simulator}
Interactions occur between the three main components of this simulator: mathematical model software, the LabVIEW-based GUI, and the rendering engine, where the former two run on two separate machines. This setup needs three LCD monitors to display the flight, the virtual cockpit, and real-time flight parameters update. The model and the engine are termed as the central component, where models are generated using Fortran with three sub-modules: aircraft dynamics, propulsion system, and atmospheric system. The rendering engine is based on C/C++ and allows many camera options that enable rotation of the virtual-vehicle during the simulation. Since the maneuverability in Micro Aerial Vehicles (MAVs) is complicated and it is necessary for an operator to master the platform, HEXAGON sports a joystick as well as a radio controller (RC). An interface RC simulator has been augmented to enable this feature and make it realistic and practical. Furthermore, the simulator training permits the users to understand complicated flights with the MP2028 autopilot system. The pilot can perform a preliminary analysis of several autopilot parameters, for instance, the proportional-integral-derivative (PID) gains, for obtaining better platform performances in real-world settings. Starting from a simple situation, it is possible to change mission parameters during the simulation~\cite{article19}.

\subsection{Test bed for a wireless network on small UAVs}
Small UAVs play an important role in applications that involve swarming and sensing. Communication within such UAVNets occurs directly or via intermediate nodes (relays) in the transit. Unmanned aerial system (UAS) is diverse with ground radios mounted either on vehicles or in sensor nodes. UAVs communicate with each other or with the ground control stations (GCS) as a distributed peer-to-peer (P2P) network. Researchers at the University of Colorado have designed a wireless network testbed using the popular IEEE 802.11b (WiFi) protocol that is used by low-cost UAVs. It provides detailed data on network parameters such as throughput, delay, range, and connectivity under different operating conditions. This data is essential for documentation and classification of network behavior among the UAVs. Furthermore, this testbed is also being used to evaluate the feasibility of Dynamic Source Routing (DSR) protocol over the 802.11b MAC. Future works may address additional routing protocols for ad-hoc networks~\cite{bekmezci2013flying,brown2004test}.

\subsection{Simbeeotic}  
\begin{figure}[h!]
\centering
\includegraphics[width=0.5\textwidth]{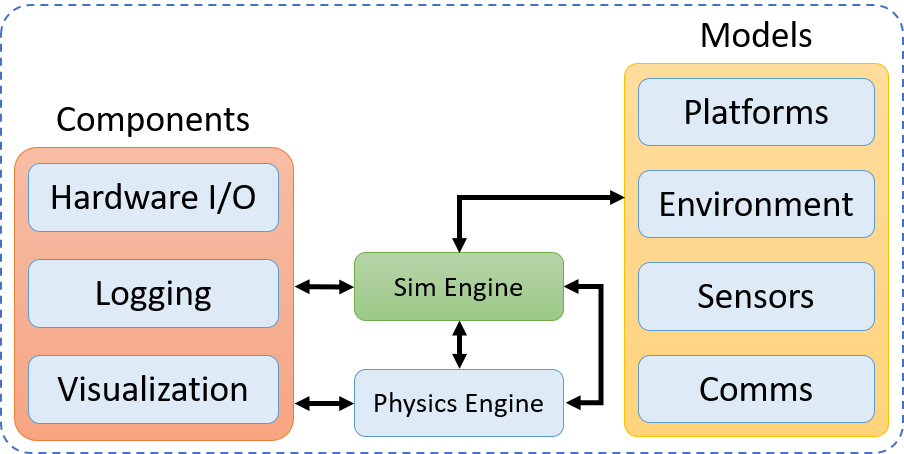}
\caption{The Simbeeotic architecture (Diagram reproduced from \cite{simbeeotic1})}
\label{fig:Simbeeotic}
\end{figure}

Simulator and Testbed for MAV Swarm Experiments (Simbeeotic) is an O/S simulator designed by Harvard University researchers as part of the RoboBees project~\cite{Autonomo71:online}. Its library provides a huge collection of MAVs that can simulate UAV swarm as well as the UAVNet communication infrastructure in a 3D virtual world~\cite{bekmezci2013flying,simbeeotic1}. Simbeeotic is an attempt to model the critical aspects of the system such as actuation, sensing, and communication. The central entity in Simbeeotic is the simulation engine which manages most vital aspects, e.g., creating the virtual world and setting models using a given configuration and model-specific initialization routine. Each module has a unique type and ID that enables the engine to locate them using an API. As shown in Figure~\ref{fig:Simbeeotic}, the model layer contains an interface for the implementation of the features related to the target domain. Users can also augment the simulator with additional components that may interact with the simulation, component engine, or the models using scheduling events. Most usable components include the 3D visualization, communication, and the Physics engine. Simbeeotic uses JBullet~\cite{JBulletJ87:online}, a Java-based physics library with six degrees of freedom (DOF) for a rigid body. JBullet provides many features that are essential for modeling MAV swarms with high fidelity~\cite{simbeeotic1}.

\subsection{SUAAVE} \label{SUAAVEu}
The Sensing Unmanned Autonomous Aerial Vehicles (SUAAVE)~\cite{suaavewp} is an interdisciplinary project sponsored by Engineering and Physical Sciences Research Council (EPSRC) under the WINES wireless networking initiative with collaboration between University College London, the University of Ulster, and the University of Oxford. The central interest of this project is the coordination and control of self-organizing, autonomous UAV swarms. Collaboratively, these light-payload quadcopter swarms sense the environment, respond to node failures, and forward reports to a base ground station. Even though the setup is not restricted to a specific scenario, it can be utilized in areas such as search \& rescue, military, and disaster management~\cite{bekmezci2013flying,article37,suap}.

The project revolves around following themes: 
\begin{enumerate}
\item Communication within a UAV swarm is a 2-step process. First, the feasibility of 802.11 is checked, then the ad-hoc or mesh networking is applied.
\item Control requires a unified approach with each UAV taking into account the resource limitations and the current state of neighbors during missions.
\item Application of Artificial Intelligence (AI) involves the use of UAVs with smart communication, command, and control for a real-world search problem.
\item Data fusion and image processing involve the utilization of multiple, possibly heterogeneous, airborne sensors and cameras for constructing and presenting accurate information for situational awareness.
\end{enumerate}

\subsection{JSBSim Flight Dynamics Model}
An O/S flight dynamics model that runs on numerous platforms such as Windows, and Linux Cygwin. Several popular frameworks such as FlightGear (O/S)~\cite{FgearFtrs}, Outerra~\cite{Outerra65:online}, BoozSimulator (O/S)~\cite{NPSPapar21:online}, and OpenEaagles (O/S)~\cite{start55:online} use this simulator. Some of its main features include a fully configurable flight control system, aerodynamic models, propulsion, landing gear arrangement, planetary rotational effects~\cite{8088163}, and engines, through changes in XML-based files. JSBSim has been used for research in motion-based simulators at the University of Naples, Italy, and the Institute of Flight System Dynamics and Institute of Aeronautics and Astronautics at RWTH Aachen University in Germany~\cite{Vogeltanz20165,berndt2004jsbsim,article39,vogeltanz2015jsbsim,article3i9}.

\subsection{FlightGear} 
The fundamental driving force behind the idea of desktop-based FlightGear project was to design an advanced flight simulator that can be beneficial for research and academia, operator training, engineering design, hobbyists (to help them identify preferred flight simulation idea), and entertainment~\cite{FgearFtrs}. Although FlightGear is primarily meant for aircraft, it can be used for simulating mini-UAVs~\cite{vogeltanz2015flightgear} and contains over 20,000 real-world airports, accurate terrains, rivers, and oceans. It has a realistic sky model, with an ability to track the current time to position the sun, moon, and stars in the sky. Networking options enable FlightGear to interact with other instances of FlightGear, GPS receivers and control modules. It is compatible with platforms such as Windows, Linux, Mac OS-X, FreeBSD, Solaris, and IRIX~\cite{Vogeltanz20165,FG,perry2004flightgear,FgearFtrs,vogeltanz2015flightgear}.

\subsection{Heli-X}
With accurate physical and mathematical models and compatible with Windows, Mac OS, and Linux, Heli-X is primarily focused on RC helicopters~\cite{Professi44:online}. However, it supports DJI Phantom~\cite{DJIPhant16:online}, a consumer drone meant for aerial photography and videography. Heli-X was designed to improve operator skills before flying the actual drone and allows users to practice in numerous re-configurable flight settings and training modes. The free version has two airports and a single training course with the default setting deliberately set by a dexterous pilot to provide a realistic touch~\cite{tr}.

\subsection{X-Plane}
A flight simulator created by Laminar Research, X-Plane runs on various platforms such as Windows, Linux, and Android~\cite{XPlane1152:online}. It enables users to create aircraft design using additional software such as Plane Maker and Airfoil Maker, and therefore, is used by a few aircraft companies. X-Plane can also create a network of its instances and communicate with UDP or TCP-based networks~\cite{ribeiro2010uav,garcia2009multi,article42}.

\subsection{AVENS}
\begin{figure}[h]
\centering
\includegraphics[width=0.70\textwidth]{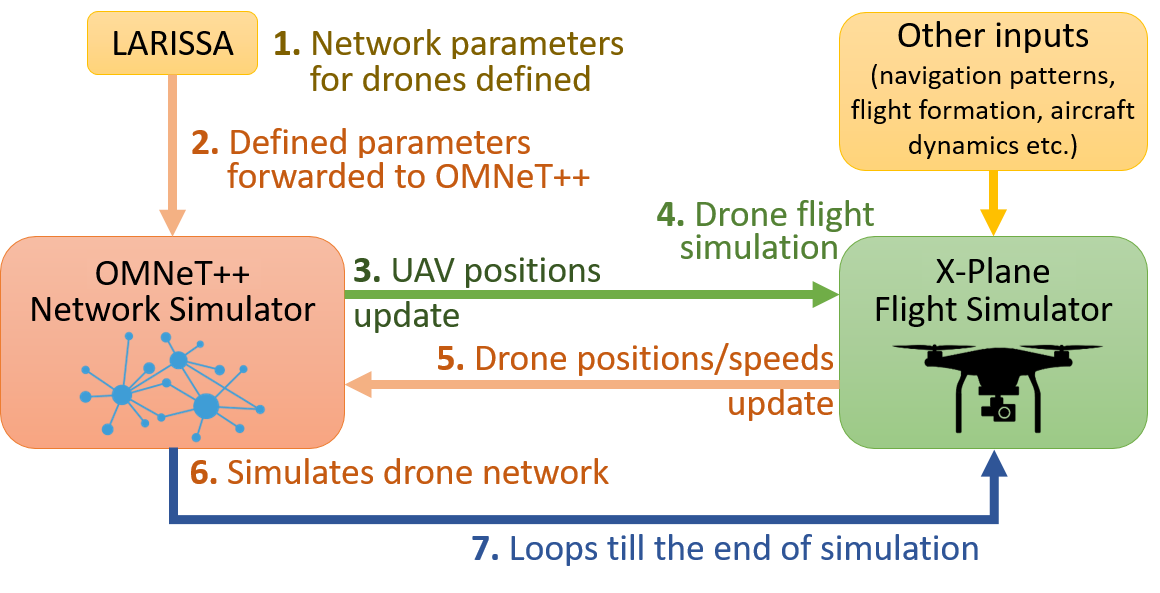}
\caption{AVENS Simulator (Diagram reproduced from \cite{marconato2017avens})}
\label{fig:figure13}
\end{figure}

Aerial Vehicle Network Simulator (AVENS) is a hybrid software-based simulator that merges X-Plane, and the OMNeT++ integrated with LARISSA (Layered architecture model for interconnection of systems in UAS)~\cite{marconato2014larissa}. Many networking protocols are developed for communication. AVENS utilizes X-Plane for controlling the flight and OMNeT++ for measuring parameters such as throughput, and packet loss. The two simulators exchange information through XML files. There is a continuous exchange of information until the simulation finishes. Detailed information exchange is shown in Figure \ref{fig:figure13}~\cite{marconato2017avens}.

\subsection{Flight Simulator}
Microsoft Flight Simulator-X~\cite{Cumming} and its previous versions are widely preferred as they feature authentic and realistic physical models of a wide variety of aircraft and UAVs such as MQ-1 Predator~\cite{FlightSi62:online} with 24000 real-world airports. The thrill of realistic flight simulation provides pilot-like experience with vivid and lively graphical landscapes in different weather conditions and seasons~\cite{article24,article26,article27,article28}.

\subsection{RAVEN}
Real-time indoor Autonomous Vehicle test Environment (RAVEN) is an indoor testbed developed by researchers at MIT for studying long endurance multi-UAV missions in a controlled environment with virtually no restrictions on flight operations. RAVEN is good at prototyping coordination and control algorithms for UAVs such as fixed and rotary wing configurations. Additionally, with RAVEN, one operator can manage a network of ten UAVs simultaneously for swarm missions~\cite{how2008real}.

\subsection{AirSim}
AirSim is an O/S drone simulator designed by Microsoft's Aerial Informatics and Robotics (AIR), primarily, to introduce it as a useful tool for AI research focusing on deep learning, computer vision, and reinforcement learning algorithms for various autonomous drones. At the outset of this idea, its simulations were limited to quadcopters. However, the AIR is intending to include other commonly used aerial robotic models. This simulator could be used to generate training data for building machine learning (ML) models. The novel feature of this simulator is the support for protocols such as Micro Air Vehicle Link (MAVLink) which aids in creating more realistic simulations~\cite{article45}.

\begin{figure}[h]
\centering
 \includegraphics[width=0.7\textwidth]{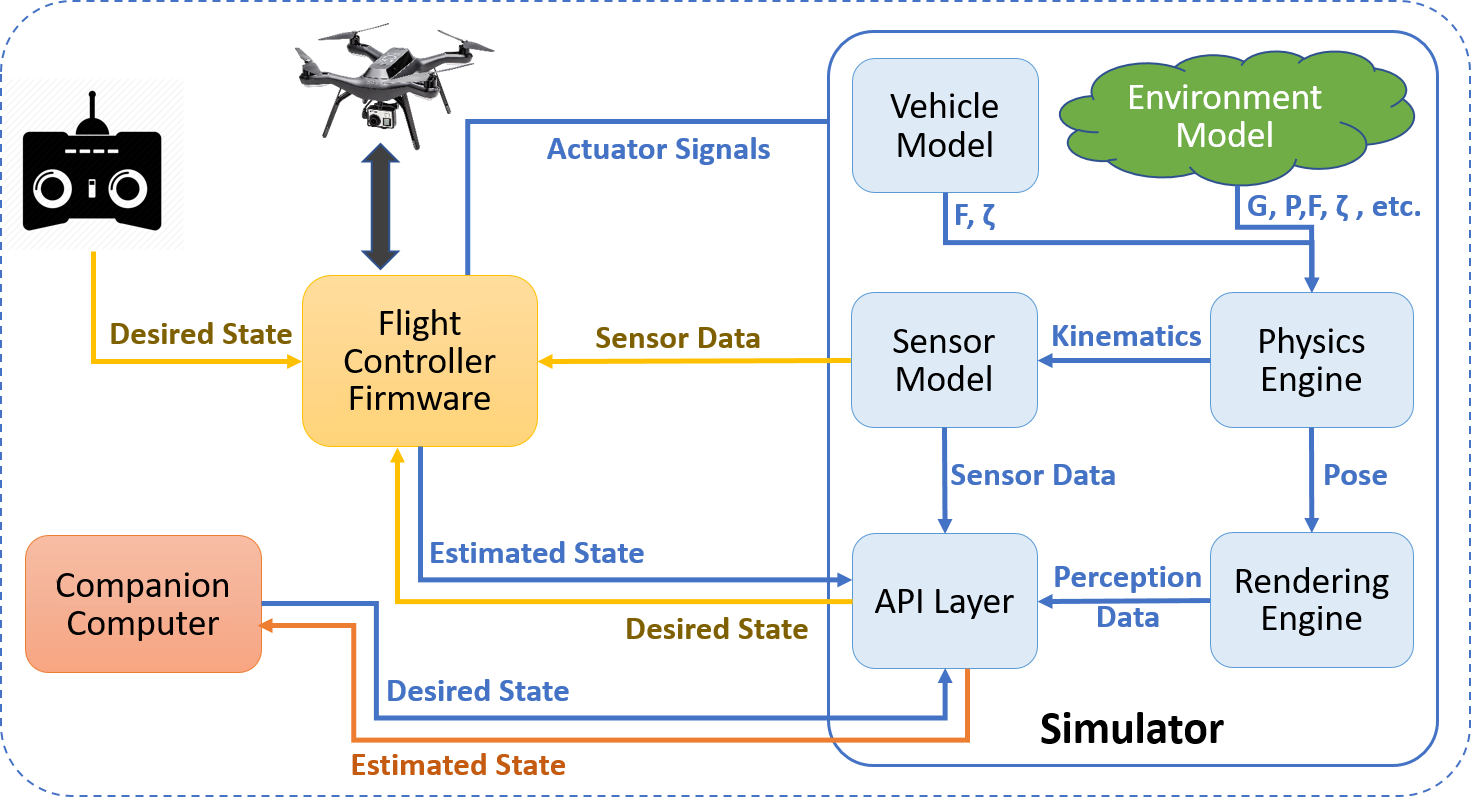}
\caption{AirSim Architectural (Diagram reproduced from \cite{airsim2017fsr})}
\label{fig:Airsim}
\end{figure}

Environment model, vehicle model, physics engine, sensor models, an API layer, rendering interface, and an interface layer form the core components of AIRSim, as shown in Figure \ref{fig:Airsim}. In general, the autonomous aerial vehicle contains the flight controller firmware such as PX4, ROSFlight, and Hackflight. The flight controller takes the desired state, and the sensor data as an input, figure out the total estimate of the current state and produces the actuator control signals to reach the desired state. For instance, in quadrotors, the user specified roll, pitch, and yaw angles are the desired state. The flight controller estimates the current angles using the data from accelerometer and gyroscope and computes the motor signals to achieve the desired angles. The simulation provides the sensor data to the flight controller which feeds the actuator signals to the vehicle model of the simulator. Vehicle model is meant for computing the forces and torques produced from the simulated actuators. For instance, in the case of quadrotors, the thrust and torques are generated by the propellers given the motor voltages are computed. Likewise, there could be other forces in play such as drag, friction, and gravity. These forces are taken as an input to the physics engine and help in computing the next kinematic state of bodies in the simulation. This kinematic state of bodies, with the environment models for gravity, air density, air pressure, magnetic field, and GPS location, is essential for the simulator sensor model. The desired state input to the flight controller can be set by the user or by a companion computer. The companion computer performs complex computations such as identifying the next desired way-point and performing Simultaneous Localization And Mapping (SLAM). It is capable of processing a large chunk of data generated from the sensors, as well as detecting collisions. The API layer is meant for shielding the companion computer from being aware of whether it is a simulation or a real-world experiment. This process helps in creating algorithms in the simulator that can later be implemented for a real vehicle without the need for any substantial modifications. The AirSim code base can be used as a plugin for the Unreal engine~\cite{airsim2017fsr,article45}.

\subsection{RotorS}
RotorS~\cite{rotors:2016} is an O/S MAV Simulator designed at Eidgen\"{o}ssische Technische Hochschule Z\"{u}rich (ETH Zurich, i.e., Swiss Federal Institute of Technology in Zurich). It includes several research-oriented multi-copter models such as the AscTec Hummingbird, Pelican, and Firefly. It also supports the addition of sensors such as camera and an inertial measurement unit (IMU) to the UAV payload. This simulator was developed for making the debugging process easier, and hence reduce crashes for real MAVs. To solve complicated tasks such as path planning, the available simulated MAVs can be used without any modifications. RotorS also include a position controller and a state estimator that works alongside the model. Gazebo plugins and Gazebo physics engine are responsible for simulating the different components present in a real MAV.

\subsection{UAVSim}
\begin{figure}
\centering
\includegraphics[width=1.0\textwidth]{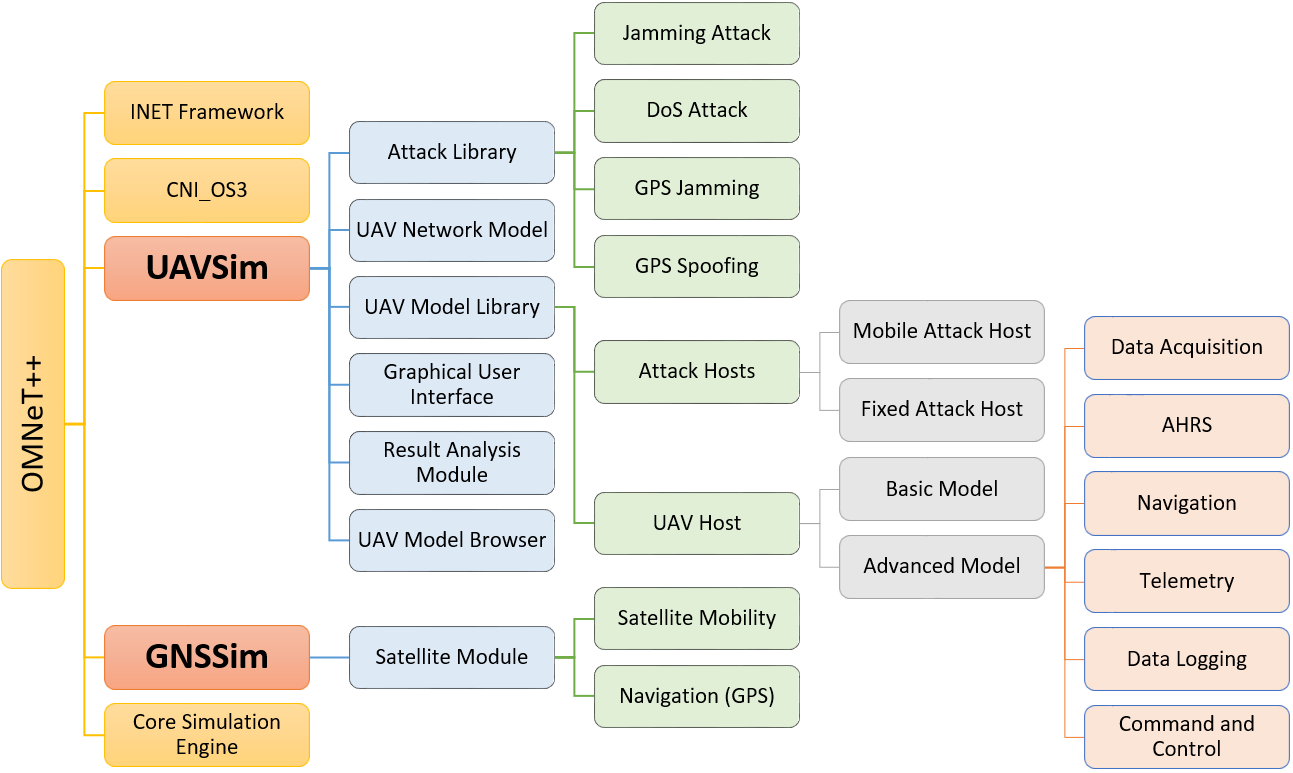}
\caption{UAVSim Modules (Reproduced from \cite{javaid2013uavsim,jahan2015gnssim,javaid2015cyber})}
\label{uavSim1}
\end{figure}

UAVSim is an OMNeT++ based testbed designed by the researchers at The University of Toledo. Its user-friendly GUI allows users to easily simulate UAV-networks by adjusting different parameters such as the number of hosts and attackers, well-defined mobility, and radio propagation models. Configurations can be set to make these simulations resemble real-world scenarios. It has a separate module for different attacks, UAV models, and a module for result and analysis (Figure~\ref{uavSim1}). The users can simulate and analyze the impact of jamming and DDoS attacks against UAVNets. This testbed can also be used to validate the communication behavior in a UAV-Network~\cite{javaid2013uavsim,javaid2015cyber,javaid2014uavnet,javaid2015single}. This simulator was further extended to develop and incorporate a GNSS simulator called GNSSim~\cite{jahan2015gnssim,jahan2015implementation}. The authors used UAVSim in combination with GNSSim to design and simulate GPS related attacks such as GPS jamming and spoofing on UAVs~\cite{javaid2017analysis}.

\begin{figure}
\centering
\includegraphics[width=0.95\textwidth]{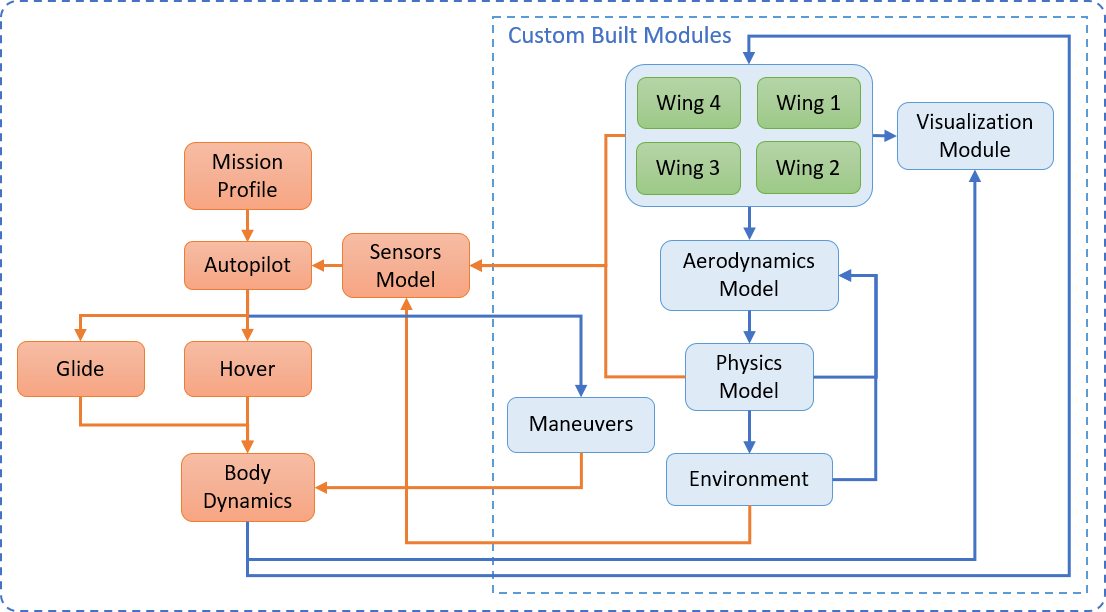}
\caption{DIMAV flight simulator (reproduced from \cite{DIMAV})}
\label{fig:DIMAV}
\end{figure}

\subsection{DIMAV}
Dragonfly Inspired Micro Air Vehicle (DIMAV) Flight Simulator~\cite{DIMAV} is aimed at analyzing the performance of a DIMAV in three different situations, namely, flight (glide or forward flight), hover, and maneuver. Figure~\ref{fig:DIMAV} provides an overview of the DIMAV flight simulator architecture. The mission profile is user-defined while the autopilot module simulates the autopilot feature of an aircraft. The autopilot is meant for execution of the mission profile by identifying the divergence between expected and the present state of the DIMAV, and for varying the wing articulation to reduce the error. The autopilot utilizes sensors such as GPS, and accelerometers to identify the present state of the drone that can be modeled in the \enquote{sensor model} block. \enquote{Physics} model defines the dynamics and behavior of wing articulation along with other physical characteristics. The equations-of-motion are evaluated using force inputs from the \enquote{aerodynamics} and the \enquote{environment} module. Modules within the dotted section such as wing dynamics, aerodynamics, and physics are related to the DIMAV maneuverability. Essential environmental factors include wind and gravity, and the aerodynamic module solves the aerodynamic forces related to selected wing motion and body dynamics.

\subsection{Vampire{$^{\textregistered}$} Simulation Suite}
\subsubsection{VAMPIRE\textsuperscript{\textregistered} PRO}
VAMPIRE\textsuperscript{\textregistered} PRO is a product of AEgis~\cite{Unmanned28:online}, which is a next-gen drone pilot training simulator. It is capable of simulating the common characteristics necessary for operator training, models emergencies accurately, and supports fixed and multi-rotor aircraft. Furthermore, the simulator provides detailed and accurate geo-typical environments, whether they are urban, sea, or mountainous terrains.

\subsubsection{Vampire{\textsuperscript{\textregistered}} UAS Embedded Training}
Another version of Vampire{\textsuperscript{\textregistered}} Simulator, features a fully embedded UAV training simulator that can emulate almost all operator functions and emergencies, especially for small drones such as Raven{\textsuperscript{\textregistered}}~\cite{RavenUAS89:online}, Wasp{\texttrademark}~\cite{WaspAEUA67:online}, and Puma{\texttrademark}~\cite{PumaAESm41:online}. It works on fielded Panasonic Toughbook and requires no additional hardware. These simulations include detailed game-quality graphics with 3D human and vehicular movements within urban areas along with graphically pleasing special effects~\cite{vampire}. AEgis is considered to be a leader in the area of drone and aerial vehicle simulation and training. Beside these simulators, Vampire{\textsuperscript\textregistered} simulation suite also include Vampire{\textsuperscript{\textregistered}} Team, Vampire{\textsuperscript{\textregistered}} Institutional Training System (ITS), and Vampire{\textsuperscript{\textregistered}} Bi-directional Advanced Trainer (BAT).

\subsection{RealFlight}
RealFlight~\cite{RealFlig37:online} is another popular flight simulator among hobbyists and pilots due to its reliability and credence. An infusion of more number of drones led to a new standalone drone edition of the simulation engine, which includes a realistic controller that duplicates the original buttons and knobs, making it realistic and practical. RealFlight includes 15 models including aerial photography, racing, and multi-copter drones. It is capable of simulating up to 8 control channels along with a dedicated flight physics engine. Apart from the free flight mode, there are additional modes with environments and flight challenges. Its versatility allows users to practice safely with a variety of drone models~\cite{dg}.

\subsection{Drone Racing League (DRL)}
DRL is focused on entertainment and fun and allows users to navigate through buildings, dockyards, old factories, and numerous other tracks meant for real-life drone racing. It supports both single and multi-player modes, with state-of-the-art maps and scenery. For race series, it has a single 25cm drone available for flight. This simulator is a training ground for novices and allows the DRL fans to race through the same courses that were introduced in 2016-17. DRL Allianz World Championship seasons were broadcast on ESPN, Sky and many other prominent sports channels across the globe. The tutorials offer missions in 50 different settings such as pitch, yaw, throttle, and creates a perfect training platform for neophytes without spending hundreds of dollars on drone practice. Users can play against anyone with the help of realistic DRL maps and accurately engineered drones. Some of the popular drones used in this simulators are Racer3, RacerX, and Nikko Air~\cite{FPR,DRLbest}.

\subsection{DroneSim Pro}
DroneSim Pro is an innovative and affordable drone simulator. It is based on realistic scenarios and accurate drone flight physics, especially the model of Phantom Vision+. This simulator is useful for both experts and novice pilots and   Windows and Mac. The graphical aspect includes detailed geological features and terrain for accuracy. New releases are expected to include additional models of drones and flight scenarios, and support for RCs~\cite{dth}.
 
\subsection{Real Drone Simulator}
Real Drone Simulator~\cite{RDSAbout66:online} is in the early phases of development and yet promising to a larger extent. It is practical in two types of environments -- (i) virtual reality, which is computer generated, and (ii) real-world using Google earth. The difficulty levels and other profile settings such as area size and wind strength can be varied. Additionally, the aerodynamics, handling, and the weather conditions are realistic. This simulator is meant for entertainment, and the developers plan to make it free. The new releases will enhance the experience with a career mode that lets users earn virtual money, which could be used to build and maintain aircraft, and race them against other online opponents. A multi-player mode allows 4-8 simultaneous players where users can control their drone with a transmitter, gamepad, keyboard or mouse connected via USB. New versions might focus on the addition of photography and aerial filming missions~\cite{tr,RDS}.

\subsection{Zephyr}
Zephyr Sim~\cite{zephyr1}, designed for pilots or drone enthusiasts willing to improve their drone flight skills, is used for industry-specific training. Zephyr currently supports several popular game-pad controllers and runs on both Windows and Mac. It also provides instructors to monitor and gauge students' progress effortlessly. Zephyr's physics models are accurately made to simulate what one can experience in real-world. This simulator was meticulously designed, where each UAV is separately modeled from the real-world information to provide a realistic experience with the primary purpose of making a seamless transition from a simulation to a real flight. Some of the drones included in Zephyr include DJI Phantom 3, DJI Inspire 1, 3DR Solo, Syma X5C, Autel X-Star, Parrot Bebop 2, and DJI Mavic Pro.

\subsection{Computational Multicopter Design}
Researchers from MIT Computer Science and Artificial Intelligence Laboratory (CSAIL) developed software for democratizing the robot design. Designing a drone is a challenging task as numerous variables should be taken into consideration such as the position of the rotor; the shape and size of the drone; controller parameter, payload, cost, and battery usage. This simulator attempts to take care of all these parameters and metrics. Once the user simulates a design, the software tells whether it will fly and allows users to experiment with a broader range of designs. One can evaluate specific unconventional designs such as the penta-copter and rabbit-shaped design. Users design drones with a database, assembling the simulated drone with preferable shapes and sizes, number of rotors, and desired payload and battery. The algorithms take a computational approach to determine the practicality of the design by measuring metrics such as torque and thrust. The goal is to find an optimized design that involves adjusting various settings and finally arriving at a trade-off between the chosen metrics. This is a real-time interactive simulator; users can change the control parameters and shapes during simulation~\cite{du2016computational,WNMit,popular,DigitalTrend,MITNewsDs}.

\begin{figure}[h]
\centering
\includegraphics[width=0.7\textwidth]{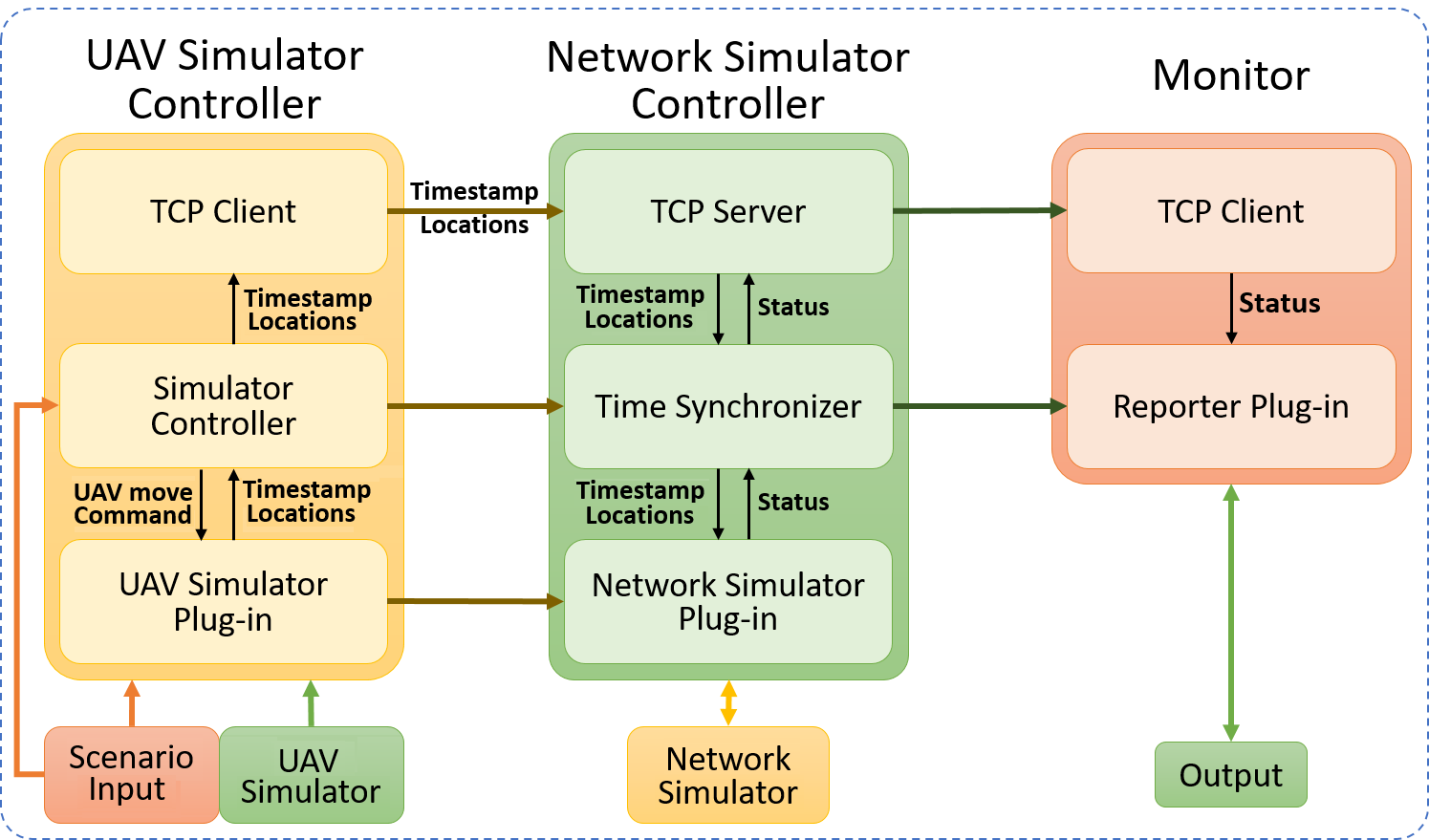}
\caption{D-MUNS simulator (Diagram reproduced from \cite{7993738})}
\label{D-MUNS}
\end{figure}

\subsection{D-MUNS}
Distributed Multiple UAVNet Simulator {D-MUNS} project~\cite{7993738} was proposed using existing simulators. For example, the RotorS can act as a medium for multiple UAV flight simulators and NS-3 as the networking simulator. D-MUNS includes three components: UAV Simulator Controller, Network Simulator Controller, and a Monitor, as shown in Figure~\ref{D-MUNS}. It was established that the simulation time and effort could be effectively reduced with the application of D-MUNS. This system is expected to contribute to the R\&D of UAVs that need simulation tests. Following is a brief description of the three components of D-MUNS:

\begin{enumerate}
\item \textbf{UAV Simulator Controller} is responsible for controlling the flow of the simulated UAVNet. It has three sub-modules: simulation controller, UAV simulator, and a TCP client. RotorS is used as a multi-UAV flight simulator that simulates various UAVs and relies on Gazebo. Once the location of UAV changes, this component directs critical information such as the current locations along with the timestamps to the Network Simulator Controller via TCP client module.
\item \textbf{Network Simulator Controller} has an interface to the actual Network Simulator, NS-3.
\item \textbf{Monitor} exports the simulation status logs as per the configuration. Through experiments, it has been verified that the real-time feature works well with time synchronizing algorithms.
\end{enumerate}

\subsection{Discussion}
UAV simulators fall in different genres, their purpose and features vary depending upon the audience they are targeting. For example, universities and research organizations use them for improving the UAVNet/FANET security, understanding drone technology, developing new UAVNet/FANET protocols, etc. While novice pilots and gamers use them for training and entertainment. Furthermore, some simulators are intended for design engineering and drone aerodynamics and efficiency enhancement. In this paper, we discussed several UAV Simulators along with their purpose, creators, network-support, public accessibility, and included UAV models. These findings are presented in Table~\ref{summary} while any missing information is represented with a hyphen (-). Some of these simulators are publicly accessible, especially if they are intended for research. Likewise, other simulators are O/S with limited features. Table~\ref{summary} also shows the simulators that have UAVNet support with collaboration and communication between drones, however, in the case of Real Drone Simulator, it refers to the same feature as multi-player mode.

\small
\begin{center}
\begin{longtable}{p{125pt} p{10pt} p{100pt} p{10pt} p{50pt} p{105pt}}
\caption{Comparison of studied drone simulators\\(O/S: Open Source, N/S: UAVNet Simulation Support), Ent: Entertainment, Train: Training}\label{summary}\\
\hline
\textbf{Simulator} & \textbf{O/S} & \textbf{Creators}& \textbf{N/S} & \textbf{Purpose} & \textbf{UAV types included } \\ \hline
HEXAGON \cite{article19} & ---  &  ---  & ---  & R\&D & MicroHawk MH-600/2000\\ \hline
Simbeeotic \cite{bekmezci2013flying}  \cite{simbeeotic1} & --- & Harvard University  &  Yes & R\&D & ---\\ \hline
SUAAVE \cite{suaavewp}  & --- & Collaborative project  &  Yes  & R\&D & ---\\ \hline
JSBSim  \cite{berndt2004jsbsim} &  Yes & JSBSim build team  & ---  & R\&D & --- \\ \hline
MS Flight Simulator X \cite{Cumming,FlightSi62:online}& No & Microsoft Studios  & --- &  Ent &  MQ-1 UAV Predator\\ \hline
FlightGear \cite{FG} \cite{perry2004flightgear} &  Yes & Developer Community & ---  &  Ent & --- \\ \hline
Heli-X \cite{tr,Creditsf48:online} & --- & Team Martin Hermann & ---  & Ent & Helicopter, DJI Phantom\\ \hline
X-Plane \cite{XPlane1152:online,ribeiro2010uav,garcia2009multi,article42}& ---  & Austin Meyer &  Yes  & R\&D, Train & ---  \\ \hline
AVENS \cite{marconato2017avens} &  Yes & Emerson A. Marconato & Yes  & R\&D & --- \\ \hline
UAVSim \cite{javaid2013uavsim,javaid2015cyber}& Yes & The University of Toledo & Yes  & R\&D & --- \\ \hline
RAVEN \cite{how2008real} & --- & MIT &  Yes  & R\&D & --- \\ \hline
Computational Multicopter Design \cite{du2016computational}&  Yes & MIT & --- & Design & ---  \\ \hline
AirSim \cite{airsim2017fsr} & Yes & MS Aerial Informatics & --- & R\&D & --- \\ \hline
RotorS \cite{rotors:2016}&  Yes & ETH Zurich & --- & R\&D & AscTec: Hummingbird,\\%\hline
\cline{1-5}
D-MUNS \cite{7993738} & --- & Korea University &  Yes  & R\&D & Pelican, Firefly\\ \hline
DIMAV \cite{DIMAV} &  Yes & Kok JM and Chahl JS & ---  & R\&D & --- \\ \hline
VAMPIRE Suite \cite{vampire} &  Yes & AEgis & ---  & Train, Ent &  Raven, Wasp, and Puma  \\ \hline
RealFlight \cite{dg} & --- & ---  & --- &Train, Ent & --- \\ \hline
Real Drone Simulator \cite{RDS} &  Yes & --- & --- & Ent & --- \\ \hline
DRL Simulator \cite{FPR}, \cite{DRLbest}  & ---  & Drone Racing League & --- &Train, Ent & Racer3, RacerX, NikkoAir\\ \hline
droneSim Pro \cite{dth} & ---  &  Jason Hershcopf & ---  & Train & Phantom Vision+\\ \hline
Zephyr Sim \cite{zephyr1,zephyr2} & --- & Little Arms Studios & --- & Train & DJI Phantom, Inspire \& Mavic Pro, 3DR Solo, Syma X5C5, Autel X-Star, Parrot Bebop 2\\ \hline
\end{longtable}
\end{center}

Table~\ref{Applications} attempts to give an insight into the possible applications of studied UAV simulators based on our understanding of these simulators. It is expected that their existing or slightly changed versions could be utilized for these and possibly, additional applications. For example, a simulator that finds its applications in mapping and surveillance could also be used by organizations such as Sunflower Labs, media industry, police departments, and fire departments. If the simulator is focused on drone swarms, it might be useful for swarm robotics for agricultural applications (SAGA), military, United Nations (UN), and humanitarian applications. Furthermore, the training and entertainment simulators could be useful for Racing Leagues, or if used for pilot training, the trained operator can fly the drone proficiently, and hence would be helpful for film-making, package delivery, precision agriculture, and organizations such as Ziplines. Research-oriented simulators can be used by security experts, universities, and data scientists. For example, UAVsim, AVENS, and D-MUNS are helpful in understanding the changing trend in FANET protocols and security loopholes. AirSim, on the other hand, generates essential data for ML, and therefore data platforms such as Kaggle, and EliteDataScience can employ it. Also, the simulators created for designing can be utilized by drone manufacturers such as DJI, AeroVironment, and Parrot for improving the aerodynamics. Such simulators can also be used to identify the critical parameters for improving the drone maneuverability. Moreover, the RotorS simulator can be enhanced for path planning for military, disaster-management, and drone farming.

\begin{center}
\begin{longtable}{p{100pt} p{80pt} p{250pt}}
\caption{Drone simulators and possible application areas \\(Ent:Entertainment, Train: Training)}\label{Applications}\\
\hline
\textbf{Simulator} & \textbf{Applications} & \textbf{Example Application Domains/Organizations}\\ \hline
HEXAGON \cite{article19} &  Survey \& Map  & Film-making \cite{Dronesin45:online}, Fire department \cite{DenverPo64:online}, Sunflower Labs \cite{Sunflowe59:online}\\ \hline
Simbeeotic \cite{bekmezci2013flying}  & MAV Swarms & Swarm for Agriculture \cite{DroneSwa7:online}, Military\cite{ThePenta99:online}\\ \hline
SUAAVE \cite{suaavewp}  & Swarm, AI, ML & Landpoint \cite{Unmanned40:online}, Humanitarian \cite{Humanita91:online}  \\ \hline
JSBSim  \cite{berndt2004jsbsim} & Flight Dynamics &  Future Drone traffic control Study, Gaming \\ \hline
MS Flight Simulator X \cite{Cumming,FlightSi62:online} & Survey, Map, Ent, Train, R\&D & Game, United Nations(UN) \cite{TheUseof9:online}, Security \\ \hline
FlightGear \cite{FG} \cite{perry2004flightgear}& R\&D, Gaming & Film-making \cite{Dronesin45:online}, Fire departments \cite{DenverPo64:online}, Security\\ \hline
Heli-X \cite{tr,Creditsf48:online} & Train, Ent & Film-making \cite{Dronesin45:online}, Fire departments \cite{DenverPo64:online}, Security, etc. \\ \hline
X-Plane \cite{XPlane1152:online,ribeiro2010uav,garcia2009multi,article42}& Design, R\&D &DJI\cite{DJIAbout79:online}, AeroVironment\cite{AeroViro14:online}, Parrot \cite{DronesOf3:online}, UAVNet Security\\ \hline 
AVENS \cite{marconato2017avens} & R\&D & UAVNet Security, Academic Curriculum  \\ \hline
UAVSim \cite{javaid2013uavsim,javaid2015cyber}& R\&D & UAVNet Security, Academic Curriculum \\ \hline
RAVEN \cite{how2008real} & UAV Swarms & Swarm for Agriculture~\cite{DroneSwa7:online}, Military \cite{ThePenta99:online}, Humanitarian \cite{Humanita91:online}, Drone Farming \cite{DroneFar0:online}, etc. \\ \hline
RotorS  \cite{rotors:2016}& Path Planning & Military \cite{ThePenta99:online}, Humanitarian UAVNet \cite{Humanita91:online}, Drone Farming \cite{DroneFar0:online}, Film-making \cite{Dronesin45:online}, Fire departments \cite{DenverPo64:online}, Sunflower Labs \cite{Sunflowe59:online}, etc. \\ \hline
DIMAV \cite{DIMAV} & Testing MAV maneuverability & MAV-Design-Industry  \\ \hline
VAMPIRE Suite \cite{vampire} &  Train, Ent & Humanitarian UAVNet \cite{Humanita91:online}, Drone Farming \cite{DroneFar0:online}, Film-making \cite{Dronesin45:online}, Fire departments \cite{DenverPo64:online}, etc.    \\ \hline
RealFlight Simulator \cite{dg} &  Train, Ent & Film-making \cite{Dronesin45:online}, Drone Racing \cite{Dronerac34:online}, Ziplines\cite{Ziplines97:online}, Precision Agriculture \cite{Agricult1:online} \\ \hline
AirSim \cite{airsim2017fsr}  &  ML Research  &  Kaggle Datasets \cite{KaggleYo77:online}, EliteDataScience \cite{EliteDat18:online}\\ \hline
Real Drone Simulator \cite{RDS} & Train, Ent & Film-making~\cite{Dronesin45:online}, Ziplines~\cite{Ziplines97:online}, Precision Agriculture \cite{Agricult1:online}, Reolink home security systems \cite{Dronesfo60:online}, Sunflower Labs \cite{Sunflowe59:online}  \\ \hline
DRL Simulator \cite{FPR}, \cite{DRLbest}  & Train, Ent & Drone Racing \cite{Gentleme35:online,Dronerac34:online}  \\ \hline
droneSim Pro \cite{dth} & Train, Ent & Package Delivery \cite{ComingSo29:online}, Drone Farming \cite{DroneFar0:online}, Film-making \cite{Dronesin45:online}, Fire departments \cite{DenverPo64:online}, Ziplines~\cite{Ziplines97:online}, Precision Agriculture \cite{Agricult1:online} \\ \hline
Zephyr Sim \cite{zephyr1,zephyr2} & Train, Ent & Humanitarian UAVNet \cite{Humanita91:online}, Drone Farming \cite{DroneFar0:online}, Film-making \cite{Dronesin45:online}, Fire departments \cite{DenverPo64:online}, Ziplines\cite{Ziplines97:online}, Precision Agriculture \cite{Agricult1:online} \\ \hline
Computational Multicopter Design \cite{du2016computational}& Design & DJI \cite{DJIAbout79:online}, AeroVironment \cite{AeroViro14:online}, Parrot \cite{DronesOf3:online} \\ \hline
D-MUNS \cite{7993738} & R\&D & UAVNet security, Academic Curriculum \\ \hline
\end{longtable}%}
\end{center}
\normalsize

\section{Challenges}
\subsection{Modern Infrastructure Compatibility}
Typically, consumer drones are remotely piloted by an operator and depend on Wi-Fi or radio connectivity for remote access. At times, such a connection may not be suitable for beyond LOS (BLOS) communication. Generally, the farthest distance that a contemporary UAV can fly is less than 5 miles. In such cases, replacing the radio control with a 4G/5G connection might eliminate the problem of control over longer distances. Although still being researched~\cite{lin2018sky,maiwald20165}, many consumer drones already support this feature~\cite{skydrone, globeuav}, and blogs that provide a DIY (Do-It-Yourself) solution~\cite{wiredcraftuav4g}. For students and hobbyists, understanding this shift in technology might be difficult, especially if it is related to convoluted networking and communication protocols. In such cases, a simulator can play an essential role in providing a lucid visual explanation of its working in real time. Nonetheless, it should be noted that many drones are limited to shorter distances due to power limitations (battery).

\subsection{Simulating Recent Drone Attacks}
As the UAV industry is still in its early stages, many vulnerabilities are unexplored~\cite{hartmann2013vulnerability,javaid2012,mansfield2013,kharchenko2018,ankitdrone}. To kick-start this process, some experiments were performed on consumer drones, and it was found, e.g., Wi-Fi data in the drone is not always encrypted. This vulnerability led to the development of Samy Kamkar's SkyJack, a software that runs on a Raspberry Pi~\cite{SamyKamk71:online} and allows the user to search for other wireless devices in monitor mode. Several drones have a manufacturer-specific block of MAC addresses, known to the public, causing the victim drone to get connected to the Wi-Fi network and de-authenticate the original RC link. Thus, SkyJack takes control and pretends to be the original owner. Similarly, Denial of Drone attack was carried by Hak5 team, where the attack was launched using an airborne Wi-Fi pineapple, and the target drone was a Parrot AR~\cite{Hak515186:online}. This code constantly looks for service-set identifiers (SSID) so that the hacker can telnet to it, kill the control program, and bring the drone down. Furthermore, backdoor software such as Maldrones can be remotely copied to a drone which is within the Wi-Fi range of another drone. Once connected, it can insert a file or kill the autopilot to take control. To the best of our knowledge, none of these attacks have been simulated so far~\cite{javaid2013uavsim,Maldrone71:online}.

\subsection{Identifying FANET Protocol Vulnerabilities}
Several telemetry and communication protocols are unsafe without in-built security features. For example, MAVLink is the header-only communication protocol used between the GCS and the UAV and contains GPS location, speed, vehicle type, and vehicle orientation information. The unencrypted payload of MAVLink protocol enables the hackers to take control of a drone that uses MAVLink. Many consumer drone autopilot systems such as PixHawk and ArduPilot use MAVLink. Since security in MAVLink is expected from transport layer protocols, a simulated UAVNet could be a possible approach for the identification of a better solution, where UAVs communicate using different transport layer protocols. The results obtained can be used to identify more secure and viable transport layer protocol for telemetry data~\cite{DroneCod13:online,MAVLinkM61:online,butcher2013securing}.
 
\subsection{UAVNet vs. Traditional Networks}
UAV simulators should also take into account the different nature of UAVNets~\cite{javaid2013uavsim,7993738}. Despite the ongoing research in this domain, it is not clear whether traditional wireless networking protocols are fit for UAVNet. Furthermore, UAVNets also differ from wireless sensor networks (WSN) or mobile ad-hoc networks (MANET). UAVNets are more complex; therefore widespread applications and application-specific requirements pose challenges and communication restrictions~\cite{article17}. There are numerous network UAV communication architectures available that rely on technologies such as GCS, satellite, multi-GCS, and FANETs. A single GCS-based network is limited to the communication range between the UAV and the GCS. A satellite-based network requires satellite communication hardware mounted on each UAV which may not be useful in severe weather conditions. Multi-GCS based network is not suitable for military missions and disaster management due to reduced trust and reliability~\cite{article18}. Unlike other types of ad-hoc networks, FANET  architecture involves the communication of few UAVs with GCS or satellites and provides real-time communication that is independent of the infrastructure. Communication links between UAVs are essential for carrying out a military mission~\cite{article18}; however, solutions that include these aspects have the range and implementation constraints. UAVNets/FANETs differ from other networks, in terms of speed, mobility, node-density, communication range, size, localization, computational power, and resource utilization~\cite{javaid2013uavsim,JAST,article18}.

\section{Future Research Direction}
\subsection{Bio-Inspired Drones }\label{Bio-Inspired Drones}
Biological evolution is a 3.8 billion-year process with which life has learned what is appropriate on this planet~\cite{Biomimicry1,eagle1,eagle2,eagle3,eagle4}. This gradual process has led to different types of flying animals to adapt themselves to the environment remarkably well. For example, in different weather conditions or environments, whether in a clustered city or a windy forest, the birds can control and adjust their flight with an excellent performance. Researchers are studying the flight-pattern of various flying animals such as birds, bats, flying snakes, and insects \cite{lentink2014bioinspired,thomas2014toward,bahlman2014wing}. People from design, computing, and other disciplines collaboratively recreate bio-inspired models to enhance the products. For example, eagles are natural drone fighters. In France, eagles are trained to swoop and punch the rogue drones out of the sky, they follow a trajectory law called "\textit{Proportional Navigation}" which is quite similar to visually guided missiles during the last seconds of their flights. Therefore, these flight patterns are studied by researchers to build predator drones~\cite{Brighton13495,thomas2014toward}. Table~\ref{Bio-inspired} summarizes some of these bio-inspired drones.

\small
\begin{center}
\begin{longtable}[h]{p{45pt}p{75pt}p{100pt}p{200pt}}
\caption{Bio-Inspired drones and their Applications} 
\label{Bio-inspired}\\ \hline
\textbf{Bio-Inspiration} & \textbf{Name } & \textbf{Study / Designers} & \textbf{Application }\\ \hline
Darwin's bark spider & SpiderMAV~\cite{spider1,spider2,zhang2017spidermav,spider3}  & Aerial Robotics Lab, Imperial College, London & Capable of spinning 25 meters of web from silk\\ \hline
Bee & Bio-inspired pollinator~\cite{pollinator1,pollinator2} & Eijiro Miyako and team & Uses high-quality cameras and GPS for accurate flower search \& tracking\\ \hline
Eye  & --- \cite{eyedrone1}  & The University of Zurich & Fast-moving, clear low-light vision, succeeded in missions where common drones failed.  \\ \hline
Dragonfly & DragonflEye~\cite{ThisGene63:online} & Charles Stark Draper Lab \& Howard Hughes Medical Institute & Collected data from human-unsafe environments\\ \hline
Bat & BatBot (B2)~\cite{Advanced96:online}&  University of Illinois & Energy-efficient, can move in 40 rotational directions due to unique musculoskeletal system\\ \hline
\end{longtable}
\end{center}
\normalsize

\begin{figure}[h]
\centering
\includegraphics[width=14cm]{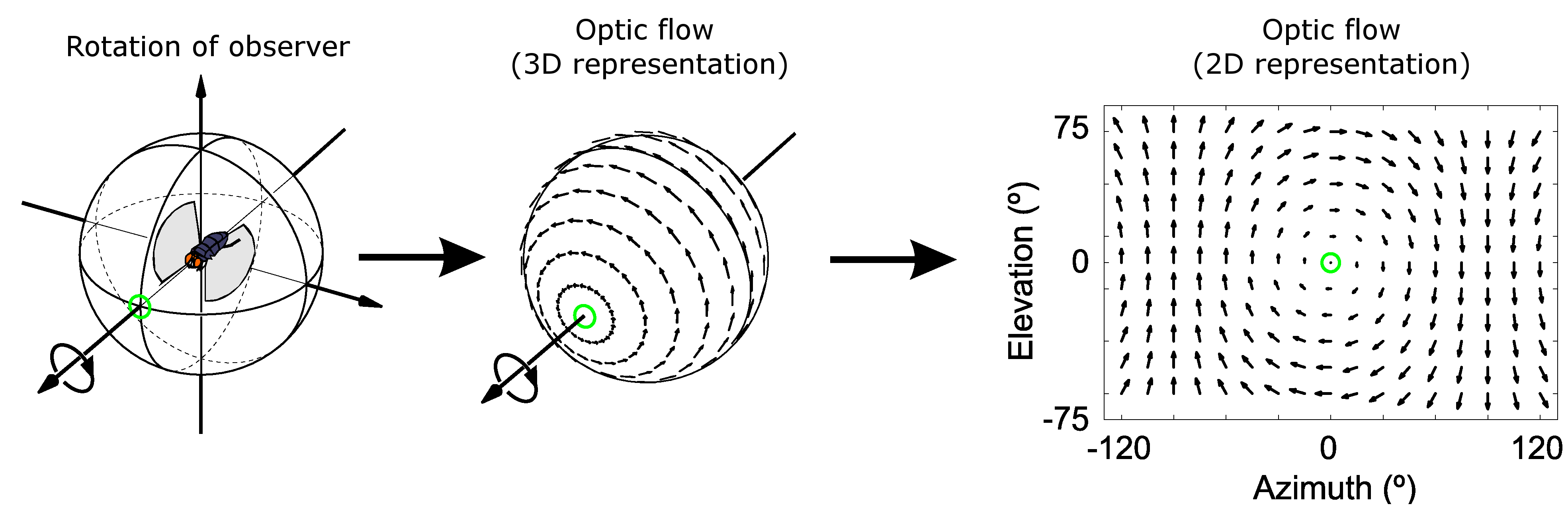}
\caption{Optic Flow in flies \cite{huston2008visuomotor}}
\label{fig:oflow}
\end{figure}

Comparatively, the flies represent the benchmark of high-performance flight among all insects, and hence drone-researchers are trying to incorporate such performance in MAVs. Fly typically relies on the calculation of \enquote{optical flow} for an effective estimation of distance~\cite{keshavan2014mu,van2014monocular,ortega2014hawkmoth}. In simple terms, optical flow is the distribution of the apparent velocities of objects within the scope of vision. Figure~\ref{fig:oflow} shows the optical flow experienced by a rotating fly. The direction and length of each arrow represent the direction and magnitude of optical flow at each location. For faster measurements of optical flow, \cite{ashraf2018low} proposes the artificial bee colony (ABC)--based block matching technique. This approach has an intelligent optical flow--based algorithm combined with Kalman filters to provide successful navigation during GPS disruptions. These features will enable the drones to fly accurately in hostile environments without relying on GPS~\cite{eyedrone1}.

\subsection{Simulating Future Drone Swarms}
Drones are getting smaller and affordable, and a swarm of such drones was employed recently at an Olympics event~\cite{olympicsdrone}. They are self-organizing and entirely independent of a leader. Swarming abilities allow the drones to perform search operations efficiently without any collisions. Swarm is very redundant and may perform well in war-zones and harsh weather conditions, e.g., the swarm may continue to proceed even after losing dozens of drones within a Network.

Continuous efforts from researchers and academia focus on improving swarm autonomy and performance. For example, the researchers at the Centre for Distributed Robotics at the University of Minnesota are working on solar-powered drones. Similarly, the University of Delft's MAV Laboratory is designing pocket-sized drone swarms. Furthermore, researchers at Harvard's Wyss Institute are developing small swarming drones that could be seen alongside insects in the near future. The RoboBee project is developing miniaturized drones with dimensions of a small paperclip weighing $\approx0.1$ gram~\cite{BBCFutur52:online}. In a swarm simulation, different settings, designs, and models could be tested and verified on lab computers. The continuous modification in the dynamics, structure, maneuverability, and other features of a single UAV is an area of fascination for hobbyists. In more intensive applications such as defense, disaster management, and mission-oriented applications, it might be critical to have a UAV swarm. Henceforth, a swarming behavior is essential to meet the continuous need for communication and reliability. Besides, we know that the swarming behavior in birds and insects is pervasive. Therefore, replicating the best of these behaviors may reveal interesting results. It's one of the crucial topics in the field of bio-mimicry. In addition, swarming in drones is an advancement towards the right direction from an application perspective~\cite{roberge2013comparison,burkle2011towards,vincent2004framework,palat2005cooperative,parunak2002digital,viragh2014flocking}

\subsection{Including New UAVs}
Many new drones, such as MAGMA drone~\cite{BAEMAGMA4:online}, are expected to have different aerodynamics. Instead of using elevators, rudders, and ailerons to control roll, pitch, and yaw, MAGMA will work with the help of a wing circulation control (WCC) and fluidic thrust vectoring (FTV). WCC takes air from the aircraft engine and blows it supersonically through the trailing edge of the UAV-Wing to control its movements. FTV, on the other hand, uses the blown air through exhausts to change the direction of the UAV. Further, speed and mobility of drones depend on their applications. Blimp drones, for example, float in the air like a balloon and are also called as Heli because they are filled with helium gas. They can capture aerial videos and function for 3 hours which is $\approx6$ times greater than a typical consumer drone. Once it runs out of battery, it does not fall and remains afloat~\cite{magma1,blimp1}.

In July 2015, Facebook announced the completion of a solar-powered drone, supposed to be tested in the UK. This flying robot had wingspan similar to that of Boeing 737 jetliner and was built to fly around the earth's stratosphere. The purpose was to provide Internet access to the most remote places of the world with lasers beams. Google also created a similar drone that crashed in New Mexico. In conclusion, to prevent such severe damage, it is better to simulate the necessary aspects of such drones before testing them in the real-world~\cite{TheFutur25:online}.

\section{Conclusion}
There has been a tangible development in the variety of available UAV simulators. However, this progress is limited to particular domains of interest. For example, most of the available UAV simulators are meant for the training, designing, and aerodynamics of a single drone. Therefore, developers invest their expertise in diversifying the simulators by including several drone models, generating accurate mathematical and aerodynamic models, and creating realistic environments. However, from a broader perspective, a good simulator demands the inclusion of many parameters and characteristics such as less complexity, easy GUI, easy implementation on generic hardware, accurate mobility models, new networking protocols, and precise mathematical and geographical details. Also, the communication and the networking is essential as UAVNets find many applications, and are continuously dependent on the exchange of data and access to different communication channels that are in turn dependent on different networking protocols. To simulate drone networking, a handful of existing network simulators are morphed into drone simulators, assuming that the drone network is similar to any other traditional ad-hoc network, however, according to recent research, the FANETs are dissimilar. Hence, a more realistic UAV Simulator demands aesthetics like any existing drone training simulator, while keeping in view the networking features of a UAVNet. A product of such kind would be promising in a variety of fields, such as academics, entertainment, research, and design engineering.

\section*{References}
\bibliography{mybibfile}

\begin{thebibliography}{100}
\expandafter\ifx\csname url\endcsname\relax
  \def\url#1{\texttt{#1}}\fi
\expandafter\ifx\csname urlprefix\endcsname\relax\def\urlprefix{URL }\fi
\expandafter\ifx\csname href\endcsname\relax
  \def\href#1#2{#2} \def\path#1{#1}\fi

\bibitem{2185196}
J.~M. Goodman, J.~Kim, S.~A. Gadsden, S.~A. Wilkerson, System and mathematical
  modeling of quadrotor dynamics, in: Unmanned Systems Technology XVII, Vol.
  9468, International Society for Optics and Photonics, 2015, p. 94680R.

\bibitem{Vijayweb}
V.~Kumar, {V}ijay {K}umar {L}ab, \url{https://www.kumarrobotics.org/},
  (Accessed: June 13, 2018).

\bibitem{javaid2015single}
A.~Y. Javaid, W.~Sun, M.~Alam, {Single and Multiple UAV Cyber-Attack Simulation
  and Performance Evaluation}, EAI Endorsed Transactions on Scalable
  Information Systems 15~(4).
\newblock \href {http://dx.doi.org/10.4108/sis.2.4.e4}
  {\path{doi:10.4108/sis.2.4.e4}}.

\bibitem{schiavullo2018ehang}
R.~Schiavullo, Ehang 184 world first self driving taxi car to flight
  autonomously at low altitude, Genesis 11 (2018) 04.

\bibitem{blackmore2014farming}
S.~Blackmore, Farming with robots 2050, in: Presentation delivered at Oxford
  Food Security Conference, 2014.

\bibitem{TheFutur99:online}
D.~Zhang, The {F}uture of {C}ommercial {D}rones | {I}dentified {T}echnologies,
  \url{https://tinyurl.com/y8sbz6z8}, (Accessed: July 19, 2018).

\bibitem{JAST}
İlker Bekmezci, E.~Şentürk, T.~Türker, {Security Issues in Flying Ad-hoc
  Network (FANETs)}, Journal of Aeronautics and Space Technologies 9~(2) (2016)
  13--21.

\bibitem{Military23:online}
S.~Smith, {M}ilitary and {C}ivilian {U}nmanned {A}erial {V}ehicles (drones),
  \url{https://tinyurl.com/y87et7ck}, (Accessed: August 14, 2018).

\bibitem{HowDrone4:online}
L.~Reich, {H}ow {D}rones are being used in {D}isaster {M}anagement? -
  {G}eoawesomeness, \url{http://geoawesomeness.com/drones-fly-rescue/},
  (Accessed: August 14, 2018).

\bibitem{TheUseof74:online}
M.~Altaweel, {T}he {U}se of {D}rones in {H}uman and {P}hysical {G}eography
  \textasciitilde{} {GIS} {L}ounge,
  \url{https://www.gislounge.com/use-drones-human-physical-geography/},
  (Accessed: August 14, 2018).

\bibitem{Dronesan5:online}
M.~A. Azevedo, {D}rones and journalism | {The Network},
  \url{https://newsroom.cisco.com/feature-content?articleId=1851973},
  (Accessed: August 14, 2018) (July 03 2017).

\bibitem{Agricult98:online}
C.~Anderson, {A}gricultural {D}rones - {MIT} {T}echnology {R}eview,
  \url{https://www.technologyreview.com/s/526491/agricultural-drones/},
  (Accessed: August 14, 2018).

\bibitem{Dronesan12:online}
S.~Long, {D}rones and {P}recision {A}griculture: {T}he {F}uture of {F}arming,
  \url{https://tinyurl.com/ydctl5kt}, (Accessed: August 14, 2018) (November
  2017).

\bibitem{DronesTr80:online}
P.~Gutierrez, {D}rones {T}ransform {A}rchaeology - {I}nside {U}nmanned
  {S}ystems,
  \url{http://insideunmannedsystems.com/drones-transform-archaeology/},
  (Accessed: August 14, 2018) (MAY 2016).

\bibitem{HBOUsedP79:online}
A.-M. Alcántara, {HBO} {U}ed {P}izza-{D}elivery {D}rones to {P}romote the
  {N}ew {S}eason of {S}ilicon {V}alley – {Adweek},
  \url{https://tinyurl.com/ybnu9zfn}, (Accessed: August 14, 2018) (March 2018).

\bibitem{Dronetax24:online}
J.~Feist, {D}rone taxi service - passenger drones - {DroneRush},
  \url{https://www.dronerush.com/drone-taxi-passenger-drones-10666/},
  (Accessed: August 14, 2018) (June 2018).

\bibitem{Vogeltanz20165}
T.~Vogeltanz, A survey of free software for the design, analysis, modelling,
  and simulation of an unmanned aerial vehicle, Archives of Computational
  Methods in Engineering 23~(3) (2016) 449--514.
\newblock \href {http://dx.doi.org/10.1007/s11831-015-9147-y}
  {\path{doi:10.1007/s11831-015-9147-y}}.

\bibitem{shrit2017new}
O.~Shrit, S.~Martin, K.~Alagha, G.~Pujolle, A new approach to realize drone
  swarm using ad-hoc network, in: Ad Hoc Networking Workshop (Med-Hoc-Net),
  2017 16th Annual Mediterranean, IEEE, 2017, pp. 1--5.

\bibitem{8088163}
C.~G.~L. Krishna, R.~R. Murphy, A review on cybersecurity vulnerabilities for
  unmanned aerial vehicles (2017) 194--199\href
  {http://dx.doi.org/10.1109/SSRR.2017.8088163}
  {\path{doi:10.1109/SSRR.2017.8088163}}.

\bibitem{hartmann2013vulnerability}
K.~Hartmann, C.~Steup, The vulnerability of uavs to cyber attacks-an approach
  to the risk assessment, in: Cyber Conflict (CyCon), 2013 5th International
  Conference on, IEEE, 2013, pp. 1--23.

\bibitem{ColinSnow}
C.~Snow, {W}hy {D}rones are the {F}uture of the {I}nternet of things - {sUAS}
  news - {T}he {B}usiness of {D}rones,
  \url{https://www.suasnews.com/2014/12/why-drones-are-the-future-of-the-internet-of-things/},
  (Accessed: July 22, 2018).

\bibitem{DBLP:journals/corr/AbdmeziemT14}
R.~Abdmeziem, D.~Tandjaoui, Internet of things: Concept, building blocks,
  applications and challenges, CoRR abs/1401.6877.
\newblock \href {http://arxiv.org/abs/1401.6877} {\path{arXiv:1401.6877}}.

\bibitem{6bestIoT0:online}
M.~Dean, {6 best IoT simulators for PC},
  \url{https://windowsreport.com/iot-simulators/}, (Accessed: June 13, 2018).

\bibitem{SimpleSo75:online}
Simplesoft's iot simulator for coap, mqtt, mqtt-sn, http/rest sensors and
  gateways., \url{http://www.simplesoft.com/SimpleIoTSimulator.html},
  (Accessed: June 13, 2018).

\bibitem{gupta2016survey}
L.~Gupta, R.~Jain, G.~Vaszkun, Survey of important issues in uav communication
  networks, IEEE Communications Surveys \& Tutorials 18~(2) (2016) 1123--1152.

\bibitem{article19}
E.~Capello, G.~Guglieri, F.~B.~Quagliotti, Uavs and simulation: an experience
  on mavs 81 (2009) 38--50.

\bibitem{allerton2009principles}
D.~Allerton, Principles of flight simulation, John Wiley \& Sons, 2009.

\bibitem{article15}
Q.~Abdullah, Welcome to geog 892 - geospatial applications of unmanned aerial
  systems | geog 892: Unmanned aerial systems,
  \url{https://www.e-education.psu.edu/geog892/node/508}, (Accessed: July 18,
  2018).

\bibitem{javaid2013uavsim}
A.~Y. Javaid, W.~Sun, M.~Alam, {UAVSim: A simulation testbed for unmanned
  aerial vehicle network cyber security analysis}, in: Globecom Workshops (GC
  Wkshps), 2013 IEEE, IEEE, 2013, pp. 1432--1436.

\bibitem{article20}
K.~Moore, {E}arly {H}istory of {F}light {S}imulation | {National Center for
  Simulation},
  \url{https://www.simulationinformation.com/education/early-history-flight-simulation},
  (Accessed: June 25, 2018).

\bibitem{article21}
R.~L~Page, R.~L~Page, Brief history of flight simulation.

\bibitem{article22}
{The history of flight simulators}, \url{https://tinyurl.com/yd8n7ssm},
  (Accessed: June 28, 2018).

\bibitem{article23}
Link trainer flight simulator - naval air station fort lauderdale museum,
  \url{https://www.nasflmuseum.com/link-trainer.html}, (Accessed: June 28,
  2018).

\bibitem{article24}
{Flight Simulator Technology Through the Years},
  \url{https://tinyurl.com/yde7cj2u}, (Accessed: June 25, 2018).

\bibitem{Cumming}
J.~Van~West, K.~Lane-Cummings, Microsoft Flight Simulator X for Pilots: Real
  World Training, John Wiley \& Sons, 2007.

\bibitem{article26}
{Free Flight Simulators Downloads}, \url{https://tinyurl.com/yblnaeqo},
  (Accessed: June 25, 2018).

\bibitem{article27}
{The Release of Microsoft Flight Simulator X},
  \url{https://flyawaysimulation.com/news/1619/}, (Accessed: June 25, 2018).

\bibitem{article28}
Microsoft flight simulator 2004 and fs over the years,
  \url{https://tinyurl.com/y9cbwe2h}, (Accessed: June 25, 2018).

\bibitem{marconato2017avens}
E.~A. Marconato, M.~Rodrigues, R.~d.~M. Pires, D.~F. Pigatto, C.~Q. Luiz~Filho,
  A.~R. Pinto, K.~R. Branco, {AVENS}-a novel flying ad hoc network simulator
  with automatic code generation for unmanned aircraft system, in: Proceedings
  of the 50th Hawaii International Conference on System Sciences, 2017.

\bibitem{tr}
C.~Ellis, {T}he best free drone simulator 2018 | {TechRadar},
  \url{https://www.techradar.com/news/the-best-free-drone-simulator},
  (Accessed: June 25, 2018).

\bibitem{jahan2015gnssim}
F.~Jahan, A.~Y. Javaid, W.~Sun, M.~Alam, {GNSSim}: An open source {GNSS}/{GPS}
  framework for uav network simulation, EAI Endorsed Transactions on Mobile
  Communications and Applications 2~(6) (2015) 1--13.

\bibitem{javaid2015cyber}
A.~Y. Javaid, Cyber security threat analysis and attack simulation for unmanned
  aerial vehicle network.

\bibitem{bekmezci2013flying}
I.~Bekmezci, O.~K. Sahingoz, {\c{S}}.~Temel, Flying ad-hoc networks (fanets): A
  survey, Ad Hoc Networks 11~(3) (2013) 1254--1270.

\bibitem{brown2004test}
T.~Brown, S.~Doshi, S.~Jadhav, J.~Himmelstein, Test bed for a wireless network
  on small uavs, in: AIAA 3rd" Unmanned Unlimited" Technical Conference,
  Workshop and Exhibit, 2004, p. 6480.

\bibitem{Autonomo71:online}
{A}utonomous {F}lying {M}icrorobots (robobees),
  \url{https://wyss.harvard.edu/technology/autonomous-flying-microrobots-robobees/},
  (Accessed: August 15, 2018).

\bibitem{simbeeotic1}
B.~Kate, J.~Waterman, K.~Dantu, M.~Welsh, Simbeeotic: A simulator and testbed
  for micro-aerial vehicle swarm experiments (2012) 49--60\href
  {http://dx.doi.org/10.1109/IPSN.2012.6920950}
  {\path{doi:10.1109/IPSN.2012.6920950}}.

\bibitem{JBulletJ87:online}
{J}bullet - {J}ava port of {B}ullet {P}hysics {L}ibrary,
  \url{http://jbullet.advel.cz./}, (Accessed: August 16, 2018).

\bibitem{suaavewp}
Suaave.org, \url{http://web4.cs.ucl.ac.uk/research/suaave/}, (Accessed: June
  28, 2018).

\bibitem{article37}
{SUAAVE}: {S}ensing {U}nmanned {A}utonomous {A}erial {V}ehicles,
  \url{https://tinyurl.com/y7wy7c24}, (Accessed: July 18, 2018).

\bibitem{suap}
S.~Cameron, S.~Hailes, S.~Julier, S.~I. McClean, G.~Parr, N.~Trigoni, M.~Ahmed,
  G.~McPhillips, R.~D. Nardi, J.~Nie, A.~Symington, L.~Teacy, S.~Waharte,
  Suaave: Combining aerial robots and wireless networking, 2010.

\bibitem{FgearFtrs}
Features – flightgear flight simulator,
  \url{https://home.flightgear.org/about/features/}, (Accessed: July 27, 2018).

\bibitem{Outerra65:online}
Outerra, \url{http://www.outerra.com/wfeatures.html}, (Accessed: August 16,
  2018).

\bibitem{NPSPapar21:online}
{NPS} - {P}aparazziuav, \url{http://wiki.paparazziuav.org/wiki/NPS}, (Accessed:
  August 16, 2018).

\bibitem{start55:online}
{M}ixed {R}eality {S}imulation {P}latform {(MIXR)},
  \url{http://www.openeaagles.org/doku.php}, (Accessed: August 16, 2018).

\bibitem{berndt2004jsbsim}
J.~Berndt, Jsbsim: An open source flight dynamics model in c++, in: AIAA
  Modeling and Simulation Technologies Conference and Exhibit, 2004, p. 4923.

\bibitem{article39}
J.~Berndt, {JSBSim Open Source Flight Dynamics Model},
  \url{http://jsbsim.sourceforge.net/index.html}, (Accessed: June 25, 2018).

\bibitem{vogeltanz2015jsbsim}
T.~Vogeltanz, R.~Ja{\v{s}}ek, Jsbsim library for flight dynamics modelling of a
  mini-uav, in: AIP Conference Proceedings, Vol. 1648, AIP Publishing, 2015, p.
  550015.

\bibitem{article3i9}
J.~Berndt, {GitHub - JSBSim-Team/jsbsim: An open source flight dynamics \&
  control software library in C++},
  \url{https://github.com/JSBSim-Team/jsbsim}, (Accessed: June 25, 2018).

\bibitem{vogeltanz2015flightgear}
T.~Vogeltanz, R.~Ja{\v{s}}ek, Flightgear application for flight simulation of a
  mini-uav, in: AIP Conference Proceedings, Vol. 1648, AIP Publishing, 2015, p.
  550014.

\bibitem{FG}
{Introduction – FlightGear Flight Simulator},
  \url{http://home.flightgear.org/about/}, (Accessed: June 25, 2018).

\bibitem{perry2004flightgear}
A.~R. Perry, The flightgear flight simulator, in: Proceedings of the USENIX
  Annual Technical Conference, 2004.

\bibitem{Professi44:online}
{P}rofessional {R/C} {H}elicopter {S}imulation,
  \url{http://www.heli-x.info/cms/}, (Accessed: August 16, 2018).

\bibitem{DJIPhant16:online}
{DJI} {P}hantom {D}rone, \url{https://www.dji.com/phantom}, (Accessed: August
  16, 2018).

\bibitem{XPlane1152:online}
{X-Plane} 11 {F}light {S}imulator | {M}ore {P}owerful. {M}ade {U}sable.,
  \url{https://www.x-plane.com/}, (Accessed: August 16, 2018).

\bibitem{ribeiro2010uav}
L.~R. Ribeiro, N.~M.~F. Oliveira, Uav autopilot controllers test platform using
  matlab/simulink and x-plane, in: Frontiers in Education Conference (FIE),
  2010 IEEE, IEEE, 2010, pp. S2H--1.

\bibitem{garcia2009multi}
R.~Garcia, L.~Barnes, Multi-uav simulator utilizing x-plane, in: Selected
  papers from the 2nd International Symposium on UAVs, Reno, Nevada, USA June
  8--10, 2009, Springer, 2009, pp. 393--406.

\bibitem{article42}
{Nuclear Projects :: Communicating with X-Plane using C\# and UDP},
  \url{http://www.nuclearprojects.com/xplane/info.shtml}, (Accessed: June 26,
  2018).

\bibitem{marconato2014larissa}
E.~A. Marconato, D.~F. Pigatto, K.~R. Branco, L.~H.~C. Branco, Larissa: Layered
  architecture model for interconnection of systems in uas, in: Unmanned
  Aircraft Systems (ICUAS), 2014 International Conference on, IEEE, 2014, pp.
  20--31.

\bibitem{FlightSi62:online}
B.~Stack, {FlightSim.Com - Review: MQ-1 UAV Predator by Abacus},
  \url{https://tinyurl.com/ybytg3bq}, (Accessed: August 16, 2018) (April 2012).

\bibitem{how2008real}
J.~P. How, B.~Behihke, A.~Frank, D.~Dale, J.~Vian, Real-time indoor autonomous
  vehicle test environment, IEEE control systems 28~(2) (2008) 51--64.

\bibitem{article45}
A.~Kapoor, {Microsoft extends AirSim to include autonomous car research -
  Microsoft Research},
  \url{https://www.microsoft.com/en-us/research/blog/autonomous-car-research/},
  (Accessed June 28, 2018).

\bibitem{airsim2017fsr}
S.~Shah, D.~Dey, C.~Lovett, A.~Kapoor, Airsim: High-fidelity visual and
  physical simulation for autonomous vehicles, in: Field and Service Robotics,
  2017.
\newblock \href {http://arxiv.org/abs/1705.05065} {\path{arXiv:1705.05065}}.

\bibitem{rotors:2016}
F.~Furrer, M.~Burri, M.~Achtelik, R.~Siegwart, Robot operating system (ros),
  Studies Comp.Intelligence Volume Number:625 The Complete Reference (Volume
  1)~(978-3-319-26052-5) (2016) Chapter 23, iSBN:978-3-319-26052-5.

\bibitem{javaid2014uavnet}
A.~Javaid, W.~Sun, M.~Alam, {UAVNet simulation in UAVSim: A performance
  evaluation and enhancement}, in: International Conference on Testbeds and
  Research Infrastructures, Springer, 2014, pp. 107--115.

\bibitem{jahan2015implementation}
F.~Jahan, {Implementation of GNSS/GPS navigation and its attacks in UAVSim
  Testbed}, Ph.D. thesis, University of Toledo (2015).

\bibitem{javaid2017analysis}
A.~Y. Javaid, F.~Jahan, W.~Sun, {Analysis of Global Positioning System
  (GPS)-based attacks and a novel GPS spoofing detection/mitigation algorithm
  for UAV simulation}, Simulation 93~(5) (2017) 427--441.

\bibitem{DIMAV}
J.~M. Kok, J.~S. Chahl, A low-cost simulation platform for flapping wing mavs,
  Proc.SPIE 9429 (2015) 9429 -- 9429 -- 7.
\newblock \href {http://dx.doi.org/10.1117/12.2084142}
  {\path{doi:10.1117/12.2084142}}.

\bibitem{Unmanned28:online}
{U}nmanned {A}ircraft systems - {V}ampire,
  \url{https://aegistg.com/vampire-pro-team/}, (Accessed: August 17, 2018).

\bibitem{RavenUAS89:online}
{R}aven {UAS} {(UAV)} - {A}erovironment, {I}nc.,
  \url{https://www.avinc.com/uas/small_uas/raven}, (Accessed: August 17, 2018).

\bibitem{WaspAEUA67:online}
{W}asp {AE} {UAS} {(UAV)} - {A}erovironment, {Inc.},
  \url{https://www.avinc.com/uas/view/wasp}, (Accessed: August 17, 2018).

\bibitem{PumaAESm41:online}
{P}uma {AE} {S}mall {UAS} {(UAV)} - {A}ero{V}ironment, {Inc.},
  \url{https://www.avinc.com/uas/small_uas/puma}, (Accessed: August 17, 2018).

\bibitem{vampire}
{U}nmanned {A}ircraft {S}ystems - {V}ampire,
  \url{https://aegistg.com/unmanned-aircraft-systems-vampire/}, (Accessed: June
  28, 2018).

\bibitem{RealFlig37:online}
{R}ealflight {RC} {F}light {S}imulator, \url{https://www.realflight.com/},
  (Accessed: August 17, 2018).

\bibitem{dg}
7 {B}est {D}rone {S}imulators and {D}rone {G}ames 2018,
  \url{http://www.droneguru.net/best-drone-simulators-and-games/}, (Accessed:
  June 28, 2018).

\bibitem{FPR}
{T}op {F}ree drone simulator 2018 | {F}pv drone reviews,
  \url{https://fpvdronereviews.com/reviews/top-free-drone-simulator-2017/},
  (Accessed: June 28, 2018).

\bibitem{DRLbest}
Z.~Dukowitz, {The World's Most Realistic Drone Flight Simulator—A
  Conversation with Ryan Gury, DRL's Head of Product, about Their Brand New
  Simulator - UAV Coach}, \url{https://tinyurl.com/y8dzoyp8}, (Accessed: June
  28, 2018).

\bibitem{dth}
{B}est {D}rone {F}light {S}imulators (and {D}rone {G}ames) of 2018,
  \url{https://www.dronethusiast.com/drone-flight-simulator/}, (Accessed: June
  28, 2018).

\bibitem{RDSAbout66:online}
{RDS} - {About us}, \url{http://www.realdronesimulator.com/about}, (Accessed:
  August 17, 2018).

\bibitem{RDS}
{R}eal {D}rone {S}imulator, \url{http://www.realdronesimulator.com/},
  (Accessed: June 28, 2018).

\bibitem{zephyr1}
{Zephyr-Sim}, \url{https://zephyr-sim.com/}, (Accessed: June 28, 2018).

\bibitem{du2016computational}
T.~Du, A.~Schulz, B.~Zhu, B.~Bickel, W.~Matusik, Computational multicopter
  design, ACM Transactions on Graphics (TOG) 35~(6) (2016) 227.

\bibitem{WNMit}
A.~Davies, {MIT} {S}tudents {B}uild {D}rone {C}ustomization and {T}esting
  {S}oftware | {WIRED},
  \url{https://www.wired.com/2017/04/nifty-mit-software-lets-design-test-drone/},
  (Accessed: June 28, 2018).

\bibitem{popular}
A.~Thompson, {N}ew {S}oftware {L}ets {Y}ou {D}esign {D}rones {L}ike {B}uilding
  {W}ith {L}egos,
  \url{https://www.popularmechanics.com/flight/drones/a24118/mit-custom-drone-design-software/},
  (Accessed: June 28, 2018).

\bibitem{DigitalTrend}
L.~Dormehl, {N}o {C}oding {E}xperience {N}eeded! {R}obot-{B}uilding {S}oftware
  {I}s {F}oolproof | {D}igital {T}rends,
  \url{https://www.digitaltrends.com/cool-tech/mit-3d-printed-robot-software/},
  (Accessed: June 28, 2018).

\bibitem{MITNewsDs}
Design your own custom drone | mit news,
  \url{http://news.mit.edu/2016/design-your-own-custom-drone-1205}, (Accessed:
  June 28, 2018).

\bibitem{7993738}
W.~G. La, S.~Park, H.~Kim, D-muns: Distributed multiple uavs network simulator
  (2017) 15--17\href {http://dx.doi.org/10.1109/ICUFN.2017.7993738}
  {\path{doi:10.1109/ICUFN.2017.7993738}}.

\bibitem{Creditsf48:online}
{C}redits for {supporting} {HELI-X}, \url{http://www.heli-x.info/cms/credits/},
  (Accessed: August 16, 2018).

\bibitem{zephyr2}
A.~Pruitt, {Hone your drone piloting skills with the Zephyr simulator -
  TechRepublic}, \url{https://tinyurl.com/y8g4byaf}, (Accessed: June 28, 2018).

\bibitem{Dronesin45:online}
S.~Varah, {D}rones in {H}ollywood: {W}hat {I}ndustry {I}s {N}ext?,
  \url{https://tinyurl.com/ybrvqfa9}, (Accessed: September 3, 2018).

\bibitem{DenverPo64:online}
N.~Phillips, {Denver Police and Fire Departments Split on Drone Use},
  \url{https://tinyurl.com/y735hq8v}, (Accessed: September 3, 2018).

\bibitem{Sunflowe59:online}
A.~Pachikov, {Sunflower Labs | Home Security Drone | About Us},
  \url{https://sunflower-labs.com/about/}, (Accessed: September 3, 2018).

\bibitem{DroneSwa7:online}
V.~T. Dario~Albani, {Drone Swarms in the Field},
  \url{https://tinyurl.com/y9ju7ut5}, (Accessed: September 3, 2018) (04 2018).

\bibitem{ThePenta99:online}
K.~Mizokami, {The Pentagon's Autonomous Swarming Drones Are the Most Unsettling
  Thing You'll See Today}, \url{https://tinyurl.com/ybq4mbg7}, (Accessed:
  September 3, 2018) (01 2017).

\bibitem{Unmanned40:online}
{Unmanned Aerial Vehicle (UAV) and Drone Based Land Surveying},
  \url{https://tinyurl.com/yccfgorn}, (Accessed: September 3, 2018).

\bibitem{Humanita91:online}
P.~Meier, {Humanitarian in the sky: drones for disaster response | Virgin},
  \url{https://tinyurl.com/ycgqokbk}, (Accessed: September 3, 2018).

\bibitem{TheUseof9:online}
K.~P. Apuuli, {The Use of Unmanned Aerial Vehicles (Drones) in United Nations
  Peacekeeping: The Case of the Democratic Republic of Congo | ASIL},
  \url{https://tinyurl.com/yajjgnmk}, (Accessed: September 7, 2018) (06 2014).

\bibitem{DJIAbout79:online}
{DJI - About Us}, \url{https://www.dji.com/company}, (Accessed: September 4,
  2018).

\bibitem{AeroViro14:online}
{AeroVironment, Inc. | Unmanned Aircraft Systems \& More},
  \url{https://www.avinc.com/}, (Accessed: September 4, 2018).

\bibitem{DronesOf3:online}
{Drones | Official Parrot® Site}, \url{https://www.parrot.com/us/drones},
  (Accessed: September 5, 2018).

\bibitem{DroneFar0:online}
T.~Luna, {Drone Farming Is Revolutionizing Agriculture},
  \url{https://tinyurl.com/yb5z292z}, (Accessed: September 3, 2018) (08 2017).

\bibitem{Dronerac34:online}
Q.~Tariq, {Drone racing taking off in new directions - Tech News | The Star
  Online}, \url{https://tinyurl.com/y9vy5svo}, (Accessed: September 4, 2018)
  (05 2018).

\bibitem{Ziplines97:online}
E.~Ackerman, {Zipline's Bigger, Faster Drones Will Deliver Blood in the United
  States This Year - IEEE Spectrum}, \url{https://tinyurl.com/yaq9npcq},
  (Accessed: September 4, 2018) (04 2018).

\bibitem{Agricult1:online}
A.~Meola, {Agricultural Drones: Precision Agriculture, Mapping \& Spraying -
  Business Insider}, \url{https://tinyurl.com/ya6cjswm}, (Accessed: September
  4, 2018) (08 2017).

\bibitem{KaggleYo77:online}
A.~Goldbloom, {Kaggle: Your Home for Data Science},
  \url{https://www.kaggle.com/}, (Accessed: September 7, 2018).

\bibitem{EliteDat18:online}
{EliteDataScience - Supercharge your data science career},
  \url{https://elitedatascience.com/}, (Accessed: September 7, 2018) (2016).

\bibitem{Dronesfo60:online}
Olivia, {Drones for Home Security and Surveillance – Must Read - Reolink
  Blog}, \url{https://reolink.com/home-security-drone/}, (Accessed: September
  4, 2018) (09 2018).

\bibitem{Gentleme35:online}
E.~Olsen, {Gentlemen, Start Your Drones - The New York Times},
  \url{https://tinyurl.com/y8vrr9wr}, (Accessed: September 4, 2018) (11 2015).

\bibitem{ComingSo29:online}
A.~Pasztor, {Coming Soon to a Front Porch Near You: Package Delivery Via Drone
  - WSJ}, \url{https://tinyurl.com/yaykorw5}, (Accessed: September 4, 2018) (03
  2018).

\bibitem{lin2018sky}
X.~Lin, V.~Yajnanarayana, S.~D. Muruganathan, S.~Gao, H.~Asplund, H.-L.
  Maattanen, M.~Bergstrom, S.~Euler, Y.-P.~E. Wang, The sky is not the limit:
  Lte for unmanned aerial vehicles, IEEE Communications Magazine 56~(4) (2018)
  204--210.

\bibitem{maiwald20165}
F.~Maiwald, A.~Schulte, 5.3-using lte-networks for uas-communication,
  Proceeding-etc2016 (2016) 166--175.

\bibitem{skydrone}
Skydrone, {Sky Drone FPV 2 LTE Modem for US/Canada},
  \url{https://tinyurl.com/ybxxznmn}, (Accessed: September 7, 2018) (2018).

\bibitem{globeuav}
G.~U. USA, {Globe UAV develops intelligent drone systems controlled across
  national borders.}, \url{http://g-uav.com/en/index.html}, (Accessed:
  September 7, 2018) (2018).

\bibitem{wiredcraftuav4g}
C.~Li, {Make your personal drone fly even farther with a 4G network
  connection}, \url{https://wiredcraft.com/blog/drone-copter-uav-4g-network},
  (Accessed: September 7, 2018) (03 2016).

\bibitem{javaid2012}
A.~Y. Javaid, W.~Sun, V.~K. Devabhaktuni, M.~Alam, Cyber security threat
  analysis and modeling of an unmanned aerial vehicle system, in: 2012 IEEE
  Conference on Technologies for Homeland Security (HST), 2012, pp. 585--590.
\newblock \href {http://dx.doi.org/10.1109/THS.2012.6459914}
  {\path{doi:10.1109/THS.2012.6459914}}.

\bibitem{mansfield2013}
K.~Mansfield, T.~Eveleigh, T.~H. Holzer, S.~Sarkani, Unmanned aerial vehicle
  smart device ground control station cyber security threat model, in: 2013
  IEEE International Conference on Technologies for Homeland Security (HST),
  2013, pp. 722--728.
\newblock \href {http://dx.doi.org/10.1109/THS.2013.6699093}
  {\path{doi:10.1109/THS.2013.6699093}}.

\bibitem{kharchenko2018}
V.~Kharchenko, V.~Torianyk, Cybersecurity of the internet of drones:
  Vulnerabilities analysis and imeca based assessment, in: 2018 IEEE 9th
  International Conference on Dependable Systems, Services and Technologies
  (DESSERT), 2018, pp. 364--369.
\newblock \href {http://dx.doi.org/10.1109/DESSERT.2018.8409160}
  {\path{doi:10.1109/DESSERT.2018.8409160}}.

\bibitem{ankitdrone}
A.~L. P.~S. Renduchintala, A.~Albehadili, A.~Y. Javaid, {Drone Forensics: A
  Preliminary Flight Log Analysis of Micro Drones}, in: Proceedings of the 2017
  International Symposium on Cyber Warfare, Cyber Defense, and Security, in
  collaboration with 2017 Conference on Computational Science and Computational
  Intelligence (CSCI-ISCW 2017), IEEE CPS, 2017, pp. 91--96.
\newblock \href {http://dx.doi.org/10.1109/CSCI.2017.15}
  {\path{doi:10.1109/CSCI.2017.15}}.

\bibitem{SamyKamk71:online}
S.~Kamkar, {SkyJack: autonomous drone hacking}, \url{https://samy.pl/skyjack/},
  (Accessed: June 30, 2018).

\bibitem{Hak515186:online}
{Hak5 1518 – Drones Hacking Drones | Technolust since 2005},
  \url{https://www.hak5.org/episodes/hak5-1518}, (Accessed: June 30, 2018).

\bibitem{Maldrone71:online}
R.~Sasi, {Maldrone the First Backdoor for drones. - Blogs - Garage4hackers
  Forum}, \url{http://garage4hackers.com/entry.php?b=3105}, (Accessed: June 30,
  2018).

\bibitem{DroneCod13:online}
{Drone Code Execution (Part 1) — \#\_shellntel},
  \url{https://tinyurl.com/ya4ek3zs}, (Accessed: June 30, 2018).

\bibitem{MAVLinkM61:online}
Mavlink micro air vehicle communication protocol - qgroundcontrol gcs,
  \url{http://qgroundcontrol.org/mavlink/start}, (Accessed: June 30, 2018).

\bibitem{butcher2013securing}
N.~Butcher, A.~Stewart, S.~Biaz, Securing the mavlink communication protocol
  for unmanned aircraft systems, Appalachian State University, Auburn
  University, USA.

\bibitem{article17}
S.~Hayat, E.~Yanmaz, R.~Muzaffar, Survey on unmanned aerial vehicle networks
  for civil applications: A communications viewpoint 18 (2016) 1--1.

\bibitem{article18}
I.~Bekmezci, I.~Sen, E.~Erkalkan, Flying ad hoc networks (fanet) test bed
  implementation (2015) 665--668\href
  {http://dx.doi.org/10.1109/RAST.2015.7208426}
  {\path{doi:10.1109/RAST.2015.7208426}}.

\bibitem{Biomimicry1}
J.~Benyus, {What Is Biomimicry? },
  \url{https://biomimicry.org/what-is-biomimicry/}, (Accessed: June 13, 2018).

\bibitem{eagle1}
T.~D. Barnett,
  \href{https://interestingengineering.com/eagles-will-no-longer-be-enlisted-as-drone-hunters}{Eagles
  Will No Longer Be Enlisted as Drone Hunters}, (Accessed May 24, 2018) (2017).
\newline\urlprefix\url{https://interestingengineering.com/eagles-will-no-longer-be-enlisted-as-drone-hunters}

\bibitem{eagle2}
H.~Detrick, \href{http://fortune.com/2017/12/05/drone-hunting-falcons/}{Can You
  Teach Drones to Hunt Each Other? Falcons May Hold the Secrets}, (Accessed May
  24, 2018) (2017).
\newline\urlprefix\url{http://fortune.com/2017/12/05/drone-hunting-falcons/}

\bibitem{eagle3}
J.~J. Roberts,
  \href{http://fortune.com/2017/02/22/drones-eagles-france/}{France Is Training
  Eagles to Kill Drones}, (Accessed May 24, 2018) (2017).
\newline\urlprefix\url{http://fortune.com/2017/02/22/drones-eagles-france/}

\bibitem{eagle4}
D.~Galeon,
  \href{https://futurism.com/air-force-studied-falcons-develop-bio-mimicking-drone-defense/}{The
  Air Force Studied Falcons to Develop a Bio-Mimicking Drone Defense},
  (Accessed May 24, 2018) (2017).
\newline\urlprefix\url{https://futurism.com/air-force-studied-falcons-develop-bio-mimicking-drone-defense/}

\bibitem{lentink2014bioinspired}
D.~Lentink, Bioinspired flight control, Bioinspiration \& biomimetics 9~(2)
  (2014) 020301--020301.

\bibitem{thomas2014toward}
J.~Thomas, G.~Loianno, J.~Polin, K.~Sreenath, V.~Kumar, Toward autonomous
  avian-inspired grasping for micro aerial vehicles, Bioinspiration \&
  biomimetics 9~(2) (2014) 025010.

\bibitem{bahlman2014wing}
J.~W. Bahlman, S.~M. Swartz, K.~S. Breuer, How wing kinematics affect power
  requirements and aerodynamic force production in a robotic bat wing,
  Bioinspiration \& biomimetics 9~(2) (2014) 025008.

\bibitem{Brighton13495}
C.~H. Brighton, A.~L.~R. Thomas, G.~K. Taylor, Terminal attack trajectories of
  peregrine falcons are described by the proportional navigation guidance law
  of missiles, Proceedings of the National Academy of Sciences 114~(51) (2017)
  13495--13500.
\newblock \href
  {http://arxiv.org/abs/http://www.pnas.org/content/114/51/13495.full.pdf}
  {\path{arXiv:http://www.pnas.org/content/114/51/13495.full.pdf}}, \href
  {http://dx.doi.org/10.1073/pnas.1714532114}
  {\path{doi:10.1073/pnas.1714532114}}.

\bibitem{spider1}
M.~Margaritoff, \href{https://tinyurl.com/ybsnorzu}{The Bio-Inspired SpiderMAV
  Drone Weaves Webs for Stability}, (Accessed May 24, 2018) (2017).
\newline\urlprefix\url{https://tinyurl.com/ybsnorzu}

\bibitem{spider2}
E.~Ackerman, Spidermav drone shoots webs for perching and stabilization - ieee
  spectrum, \url{https://tinyurl.com/ycqmsqgv}, (Accessed: June 29, 2018).

\bibitem{zhang2017spidermav}
K.~Zhang, P.~Chermprayong, T.~Alhinai, R.~Siddall, M.~Kovac, Spidermav:
  Perching and stabilizing micro aerial vehicles with bio-inspired tensile
  anchoring systems, in: Intelligent Robots and Systems (IROS), 2017 IEEE/RSJ
  International Conference on, IEEE, 2017, pp. 6849--6854.

\bibitem{spider3}
M.~Margaritoff, \href{https://tinyurl.com/ybsnorzu}{The Bio-Inspired SpiderMAV
  Drone Weaves Webs for Stability}, (Accessed May 24, 2018) (2017).
\newline\urlprefix\url{https://tinyurl.com/ybsnorzu}

\bibitem{pollinator1}
C.~Ponti, Rise of the robot bees: Tiny drones turned into artificial
  pollinators : The salt : Npr, \url{https://tinyurl.com/y7nqsnft}, (Accessed:
  July 23, 2018).

\bibitem{pollinator2}
J.~Kite-Powell, {See How This Bio-Inspired Drone Can Artificially Pollinate A
  Flower}, \url{https://tinyurl.com/y98halze}, (Accessed: June 25, 2018).

\bibitem{eyedrone1}
G.~Nichols, {Bio-inspired retina allows drones to almost see in the dark with
  no motion blur | ZDNet}, \url{https://tinyurl.com/y7dmpk64}, (Accessed: June
  25, 2018).

\bibitem{ThisGene63:online}
{This Genetically-Modified Cyborg Dragonfly Is the Tiniest Drone},
  \url{https://tinyurl.com/y9tz3g4z}, (Accessed: September 8, 2018).

\bibitem{Advanced96:online}
R.~Kubetz, {Advanced robotic bat's flight characteristics simulates the real
  thing - Illinois Engineering},
  \url{https://engineering.illinois.edu/news/article/21209}, (Accessed:
  September 8, 2018) (02 2017).

\bibitem{keshavan2014mu}
J.~Keshavan, G.~Gremillion, H.~Escobar-Alvarez, J.~Humbert, A $\mu$
  analysis-based, controller-synthesis framework for robust bioinspired visual
  navigation in less-structured environments, Bioinspiration \& biomimetics
  9~(2) (2014) 025011.

\bibitem{van2014monocular}
F.~Van~Breugel, K.~Morgansen, M.~H. Dickinson, Monocular distance estimation
  from optic flow during active landing maneuvers, Bioinspiration \&
  biomimetics 9~(2) (2014) 025002.

\bibitem{ortega2014hawkmoth}
V.~M. Ortega-Jimenez, R.~Mittal, T.~L. Hedrick, Hawkmoth flight performance in
  tornado-like whirlwind vortices, Bioinspiration \& biomimetics 9~(2) (2014)
  025003.

\bibitem{ashraf2018low}
S.~Ashraf, P.~Aggarwal, P.~Damacharla, H.~Wang, A.~Y. Javaid, V.~Devabhaktuni,
  A low-cost solution for unmanned aerial vehicle navigation in a global
  positioning system-denied environment, International Journal of Distributed
  Sensor Networks 14~(6) (2018) 1550147718781750.

\bibitem{huston2008visuomotor}
S.~J. Huston, H.~G. Krapp, Visuomotor transformation in the fly gaze
  stabilization system, PLoS biology 6~(7) (2008) e173.

\bibitem{olympicsdrone}
B.~Barrett, {Inside the Olympics opening ceremony world-record drone show},
  \url{https://www.wired.com/story/olympics-opening-ceremony-drone-show/},
  (Accessed: September 7, 2018) (February 2018).

\bibitem{BBCFutur52:online}
D.~Hambling, {BBC} - {F}uture - {T}he {n}ext {e}ra of {d}rones will be defined
  by 'swarms', \url{https://tinyurl.com/ms5f25z}, (Accessed: July 19, 2018).

\bibitem{roberge2013comparison}
V.~Roberge, M.~Tarbouchi, G.~Labont{\'e}, Comparison of parallel genetic
  algorithm and particle swarm optimization for real-time uav path planning,
  IEEE Transactions on Industrial Informatics 9~(1) (2013) 132--141.

\bibitem{burkle2011towards}
A.~B{\"u}rkle, F.~Segor, M.~Kollmann, Towards autonomous micro uav swarms,
  Journal of intelligent \& robotic systems 61~(1-4) (2011) 339--353.

\bibitem{vincent2004framework}
P.~Vincent, I.~Rubin, A framework and analysis for cooperative search using uav
  swarms, in: Proceedings of the 2004 ACM symposium on Applied computing, ACM,
  2004, pp. 79--86.

\bibitem{palat2005cooperative}
R.~C. Palat, A.~Annamalau, J.~Reed, Cooperative relaying for ad-hoc ground
  networks using swarm uavs, in: Military Communications Conference, 2005.
  MILCOM 2005. IEEE, IEEE, 2005, pp. 1588--1594.

\bibitem{parunak2002digital}
H.~V. Parunak, M.~Purcell, R.~O'Connell, Digital pheromones for autonomous
  coordination of swarming uav's, in: 1st UAV Conference, 2002, p. 3446.

\bibitem{viragh2014flocking}
C.~Vir{\'a}gh, G.~V{\'a}s{\'a}rhelyi, N.~Tarcai, T.~Sz{\"o}r{\'e}nyi,
  G.~Somorjai, T.~Nepusz, T.~Vicsek, Flocking algorithm for autonomous flying
  robots, Bioinspiration \& biomimetics 9~(2) (2014) 025012.

\bibitem{BAEMAGMA4:online}
B.~Brimelow, {BAE} {MAGMA} drone uses jets of air to maneuver - {B}usiness
  {I}nsider, \url{https://tinyurl.com/ycfjklmk}, (Accessed: August 17, 2018)
  (2017 December).

\bibitem{magma1}
J.~Trevithick, {BAE Systems Wants Its MAGMA Drone to Maneuver Using Only
  Supersonic Blasts of Air - The Drive}, \url{https://tinyurl.com/y7ewxppq},
  (Accessed: June 28, 2018).

\bibitem{blimp1}
C.~Dillow, {This Hybrid Drone-Blimp is a Crowd-Friendly Aerial Advertising
  Vehicle | Fortune},
  \url{http://fortune.com/2016/03/17/drone-blimp-advertising/}, (Accessed: June
  28, 2018).

\bibitem{TheFutur25:online}
E.~Gent, {T}he {F}uture of {D}rones: {U}ncertain, {P}romising and {P}retty
  {A}wesome, \url{https://tinyurl.com/ycvbhwpf}, (Accessed: July 19, 2018).

\end{thebibliography}

\end{document}